%% file: main.tex
\documentclass{article}
\usepackage{iclr2027_conference,times}
\usepackage[utf8]{inputenc}
\usepackage[T1]{fontenc}
\usepackage{hyperref}
\usepackage{url}
\usepackage{booktabs,multirow}
\usepackage{amsmath,amsfonts}
\usepackage{nicefrac,microtype}
\usepackage{xcolor,graphicx,enumitem,xspace,colortbl}
\usepackage[most]{tcolorbox}
\usepackage{wrapfig}

\newcommand{\benchmark}{{\sc DataVista}\xspace}
\newif\ifshowrevisions
\showrevisionsfalse
\newcommand{\rev}[1]{%
  \ifshowrevisions\textcolor{blue}{#1}\else#1\fi}
\definecolor{heatA}{RGB}{226,176,131}
\definecolor{humanrow}{RGB}{235,243,242}
\newcommand{\cmark}[1]{%
  \ifdim #1pt > 60pt \cellcolor{heatA!45}\textbf{#1}%
  \else\ifdim #1pt > 50pt \cellcolor{heatA!30}#1%
  \else\ifdim #1pt > 40pt \cellcolor{heatA!18}#1%
  \else\ifdim #1pt > 30pt \cellcolor{heatA!8}#1%
  \else {\color{black!50}#1}%
  \fi\fi\fi\fi
}
\definecolor{backgray}{gray}{0.95}
\definecolor{framegray}{gray}{0.45}
\newtcolorbox{examplebox}[1]{
  width=\linewidth,breakable,colback=backgray,colframe=framegray,
  boxrule=0.5pt,leftrule=2pt,toprule=1pt,bottomrule=2pt,rightrule=2pt,
  arc=0pt,title={\textbf{#1}},fonttitle=\bfseries,
  before skip=0.5\baselineskip,after skip=0.5\baselineskip
}

\title{DataVista: Diagnosing Multimodal LLMs on Data Video Understanding}

\author{%
 \textbf{Yupeng Xie}\textsuperscript{1},
 \textbf{Zhenyang Wang}\textsuperscript{1},
 \textbf{Jiayi Zhu}\textsuperscript{1},
 \textbf{Yinghao Tang}\textsuperscript{2},
 \textbf{Zhouan Shen}\textsuperscript{1},\\
 \textbf{Yiyu Chen}\textsuperscript{1},
 \textbf{Yuyu Luo}\textsuperscript{1}\thanks{Corresponding author: Yuyu Luo (E-mail: yuyuluo@hkust-gz.edu.cn)}\\[0.4em]
 \textsuperscript{1}The Hong Kong University of Science and Technology (Guangzhou)\\
 \textsuperscript{2}Zhejiang University
}

\iclrfinalcopy

\begin{document}
\maketitle

\begin{abstract}
Data video is a media form that integrates data visualization with video narrative, widely adopted in news reporting and business analysis.
\rev{Compared with general video understanding, data video understanding places greater emphasis on accurately reading data from animated charts, integrating evidence across charts and time, and understanding how narrative organization and visual design communicate information.}
\rev{Yet existing benchmarks target either general videos or static charts, and data video understanding has not been systematically evaluated.}
We present \benchmark, the first benchmark for data video understanding, containing 961 real-world data videos and 6{,}775 evaluation questions organized under a three-level progressive capability framework \rev{(data perception, temporal reasoning, narrative understanding)} with 10 fine-grained question types across five topic domains.
\rev{Systematic evaluation of 19 mainstream MLLMs shows that the best-performing model, Gemini-3.1-Pro, achieves 70.0\% overall accuracy, still far below human expert performance, with models performing worst {on Causal Reasoning and Narrative Structure}.}
\rev{Increasing frame counts and adding subtitles mainly benefit data perception and temporal reasoning, with limited gains in narrative understanding.}
\rev{Further analysis of model responses identifies typical failure modes in chart reading, evidence judgment, and instruction understanding.}
The benchmark is available at \url{https://github.com/HKUSTDial/DataVista}.
\end{abstract}

\input{src/01_introduction.tex}
\input{src/02_related_work.tex}
\input{src/03_benchmark_construction.tex}
\input{src/04_experiments.tex}
\input{src/05_conclusion.tex}

\input{src/ai_use_statement.tex}
\input{src/ethics_statement.tex}
\input{src/reproducibility_statement.tex}

\bibliographystyle{iclr2027_conference}
\bibliography{references}

\clearpage
\appendix
\input{src/appendix-a-Capability_Hierarchy}
\input{src/appendix-b-dataset_statistics}
\input{src/appendix-c-case-studies}
\input{src/appendix-d-evaluation-details}
\input{src/appendix-e-data_contamination_analysis}
\input{src/appendix-f-limitation}
\end{document}

%% file: src/01_introduction.tex
\section{Introduction}

Data videos are a media form that integrates data visualization with video narrative, conveying data-driven stories over time through animation, narration, and scene transitions~\citep{amini2015understanding,shi2021communicating,shen2025empiricaldatavideo}.
In recent years, organizations such as Bloomberg Graphics, NYT The Upshot, and Vox have produced a growing volume of such content, establishing data videos as an important medium for data journalism, business reporting, and science communication~\citep{DBLP:journals/tvcg/YangXLLGSQC22, cheng2022narration}.
Data videos provide a challenging setting for evaluating the combined capabilities of MLLMs in dynamic visual perception, cross-temporal reasoning, and narrative understanding.

\begin{figure*}[t]
  \vspace{-1em}
  \centering
  \includegraphics[width=\textwidth]{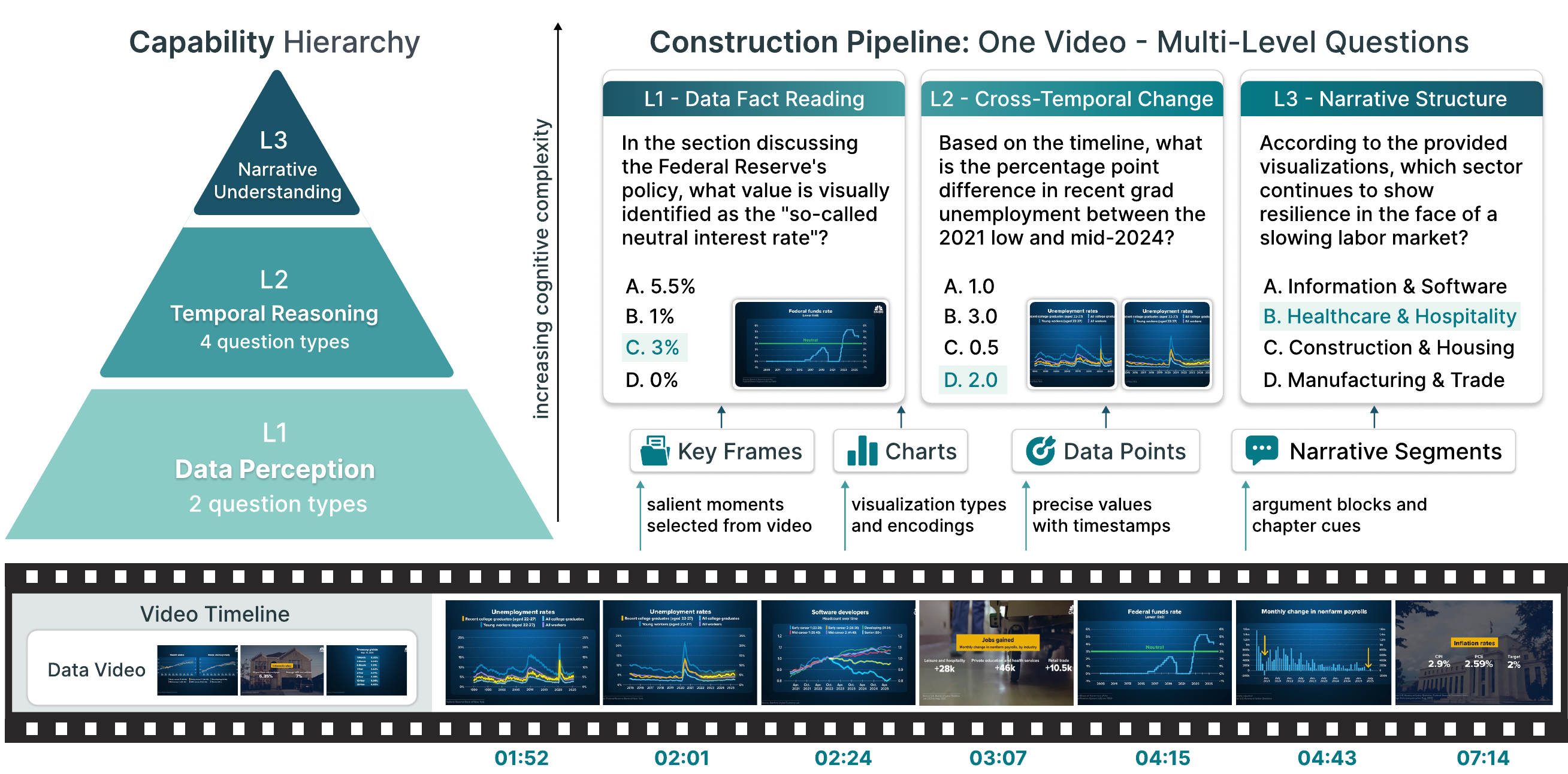}
  \vspace{-2em}
  \caption{Three-level capability hierarchy and construction pipeline of \benchmark.}
  \label{fig:dvbench_main}
  \vspace{-1.5em}
\end{figure*}

However, understanding data videos goes well beyond what general video perception covers.
As shown in Figure~\ref{fig:dvbench_main}, charts are presented within continuous videos rather than provided as isolated images~\citep{thompson2020understanding}; key information is often distributed across multiple time points~\citep{li2026dataspace}; and \rev{visual presentation also organizes narratives and communicates information~\citep{tang2026vividoc,li2024coinsight}.}
General video understanding benchmarks (Video-MME~\citep{fu2024videomme}, MVBench~\citep{li2024mvbench}) focus on perception and reasoning over general video content, whereas static visualization benchmarks (ChartQA~\citep{masry2022chartqa}, VisJudge-Bench~\citep{xie2026visjudgebench}) primarily address the understanding and evaluation of static charts. Jointly assessing evidence localization, cross-chart integration, and narrative understanding within continuous videos still lacks systematic benchmark support.

To fill this gap, we present \benchmark, the first benchmark for data video understanding.
Drawing on Bloom's cognitive taxonomy~\citep{bloom1956taxonomy}, visualization literacy assessment frameworks~\citep{DBLP:journals/tvcg/BoyRBF14, DBLP:journals/tvcg/LeeKK17}, and design spaces for data storytelling~\citep{segel2010narrative}, we systematically analyze the core abilities required for data video understanding and distill three progressive capability dimensions: data perception (accurately reading values and identifying visual elements from video frames), temporal reasoning (integrating information across time points and across charts), and {narrative understanding} (grasping the creator's visual design intent and argument organization).
\rev{These three levels form a progressive evaluation framework comprising 10 fine-grained question types.}

\benchmark contains 961 professionally produced data videos from real-world sources, spanning five topic domains (Economy, Society, Science, Politics, and Culture), paired with 6{,}775 evaluation questions, all in multiple-choice format including both single- and multi-select items.
Using \benchmark, we systematically evaluate 19 mainstream MLLMs. The results show that even the most advanced model, Gemini-3.1-Pro (70.0\% accuracy), lags significantly behind human performance, with {Causal Reasoning and Narrative Structure} being the weakest question types across all models. Furthermore, the benefits of increasing frame counts and adding subtitles vary across capability levels, with relatively limited gains in narrative understanding. Error analysis further reveals typical failure modes in chart reading, evidence judgment, and instruction understanding.

Our contributions are summarized as follows:
\begin{itemize}
    \item We present \benchmark, containing 961 real-world data videos and 6{,}775 evaluation questions, as the first dedicated benchmark for data video understanding.
    \item \rev{We design a three-level progressive capability framework with 10 fine-grained question types to systematically evaluate abilities ranging from chart reading and cross-chart evidence integration to understanding narrative and visual design.}
    \item We systematically evaluate 19 mainstream MLLMs and analyze how performance differs across capability levels, input settings, and error modes.
\end{itemize}

%% file: src/02_related_work.tex
\section{Related Work}

\paragraph{Video understanding benchmarks.}
\rev{Existing benchmarks evaluate multimodal understanding, temporal reasoning, and long-video comprehension.
Video-MME~\citep{fu2024videomme} and Video-MME-v2~\citep{fu2025videommev2} cover diverse video content and input modalities; MVBench~\citep{li2024mvbench}, NExT-QA~\citep{xiao2021nextqa}, and EgoSchema~\citep{mangalam2023egoschema} examine temporal, causal, and long-range reasoning.
LongVideoBench~\citep{wu2024longvideobench}, LVBench~\citep{wang2024lvbench}, MLVU~\citep{zhou2024mlvu}, and Neptune~\citep{wang2024neptune} further assess understanding over long videos.
Specialized benchmarks address human-centric perception and cognition~\citep{chen2025humanvideomme}, video aesthetics~\citep{he2026videoaesbench}, and driving-scene understanding~\citep{zhang2025dvbench}.
In contrast, \textsc{DataVista} focuses on quantitative evidence in data videos, evaluating how models read dynamically presented chart information, integrate evidence across charts, and understand how this evidence supports the video's claims.}

\paragraph{Chart understanding benchmarks.}
\rev{Chart understanding benchmarks have progressed from visual question answering on individual charts to multi-chart and higher-order reasoning.
ChartQA~\citep{masry2022chartqa}, PlotQA~\citep{methani2020plotqa}, and ChartInsights~\citep{wu2024chartinsights} assess visual and logical reasoning over static charts; ChartQAPro~\citep{ahmed2025chartqapro} and RWA-ChartQA~\citep{li2025realworldchartqa} introduce more diverse, real-world-oriented questions; and MultiChartQA~\citep{zhu2025multichartqa} extends to multi-hop reasoning across chart collections.
Adjacent tasks cover chart summarization, comparison, and retrieval (Chart-to-Text~\citep{kantharaj2022charttotext}, ChartCards~\citep{wu2025chartcards}, ChartDiff~\citep{islam2026chartdiff}, LineNet~\citep{luo2020linenet}) and visualization quality assessment (VisJudge-Bench~\citep{xie2026visjudgebench} and IGenBench~\citep{wang2025igenbench}).
Across all these settings, charts are either isolated images or pre-assembled static collections.
\textsc{DataVista} moves evaluation to dynamically presented charts in narrated video: models must read values from video frames, integrate evidence as it accumulates over time, and reason about how narrative structure and visual design convey the video's argument.}

\paragraph{Visualization authoring.}
\rev{Visualization authoring research covers the full pipeline from chart generation to recommendation and revision~\citep{qin2020vizsurvey,luo2024intelligent,luo2022tvcgnli}.
Natural language-driven approaches span generation benchmarks and models (nvBench~\citep{luo2021nvbench}, nvBench~2.0~\citep{luo2025nvbench2}, DeepVIS~\citep{shuai2025deepvis}, and ncNet~\citep{luo2018ncnet}).
Automatic generation and recommendation systems (DeepEye~\citep{luo2018deepeye,luo2018deepeyekeyword,luo2022steerable}, HAIChart~\citep{xie2024haichart}, chart-plot~\citep{tang2026demonstrating}, and sketch-plot~\citep{tang2026sketch}) and chart annotation and defect debugging~\citep{chen2025chartmark,shen2026debugging} address the correctness and quality of produced artifacts.
The artifacts these systems produce form the building blocks of data videos, yet whether multimodal models can understand how charts function within dynamic, narrated sequences remains unevaluated.
\textsc{DataVista} fills that gap: it examines whether models can read, integrate, and interpret the charts and narratives that authoring systems produce.}

\paragraph{Data video narrative.}
Data videos use animated visualizations and narration to communicate data stories.
\rev{Research has progressed from narrative visualization genres~\citep{segel2010narrative} and video clip structure and viewer engagement~\citep{amini2017dataclips,amini2018hooked} to the narrative role of animation~\citep{shi2021communicating}, story structure~\citep{wei2024herosjourney}, and narration--animation coordination~\citep{cheng2022narration,shen2024dataplaywright}.}
Systems such as Data Player~\citep{shen2024dataplayer}, DataMagic~\citep{xie2026datamagic,xie2026datamagicvis}, and DeepEye~\citep{li2026deepeye,li2026deepeye_workflow} support automated data video generation.
\rev{These studies focus on designing and generating data stories~\citep{zhu2025dataagents,luo2026agenticva}, while \textsc{DataVista} examines whether models understand the data relationships and narrative expression within these stories, extending data video research from content creation to understanding evaluation.}

%% file: src/03_benchmark_construction.tex
\section{\benchmark: Design and Construction}
\label{sec:construction}

\begin{figure*}[t!]
  \vspace{-1em}
  \centering
  \includegraphics[width=\textwidth]{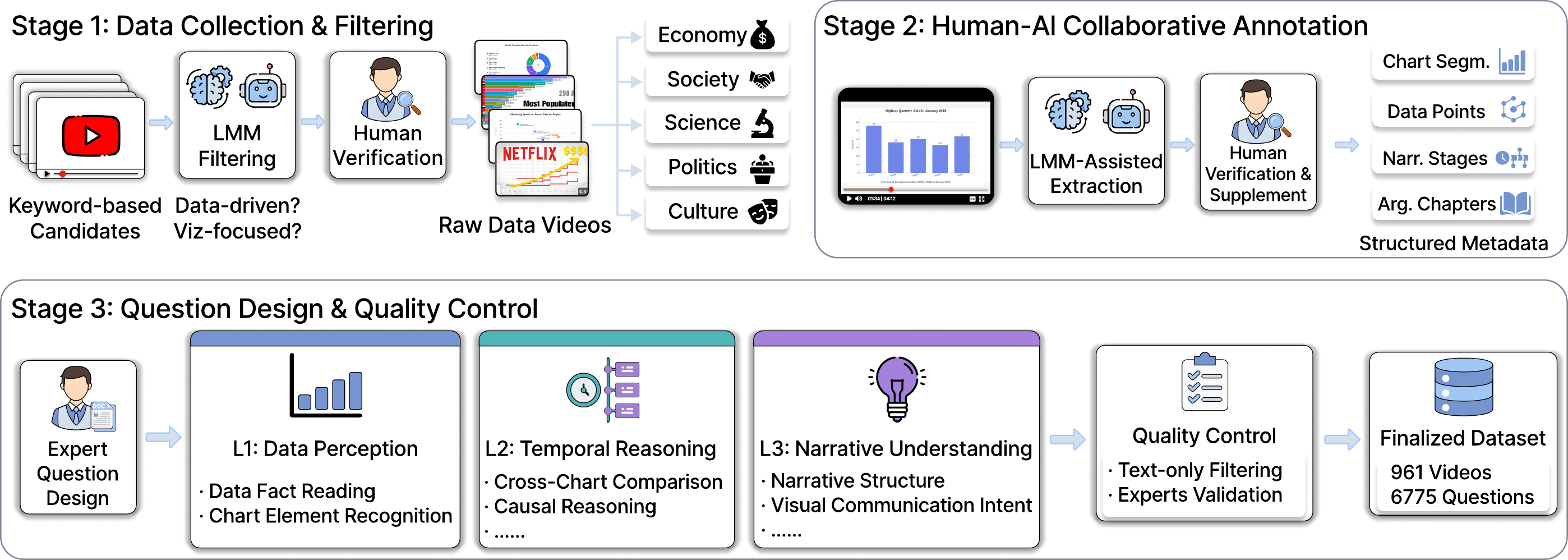}
  \vspace{-1.5em}
  \caption{Overview of the \benchmark construction process, including data collection, human-AI collaborative annotation, and question design.}
  \label{fig:benchmark_process}
  \vspace{-1.5em}
\end{figure*}

Figure~\ref{fig:benchmark_process} shows the three-stage construction pipeline of \benchmark: (1) video collection and filtering, (2) capability hierarchy and question design, and (3) expert annotation and quality control.

\subsection{Video Collection and Preprocessing}
\label{sec:collection}

\paragraph{Sources.}
Drawing on prior collection practices~\citep{DBLP:journals/tvcg/YangXLLGSQC22,fu2025videommev2}, we collect 5{,}604 candidate videos from YouTube through keyword search and channel-level scraping, covering professional media, data journalism, educational, and independent creator channels. Representative channels and search keywords are listed in Appendix~\ref{app:collection}.

\begin{figure*}[t]
  \centering
  \vspace{-1em}
  \includegraphics[width=\textwidth]{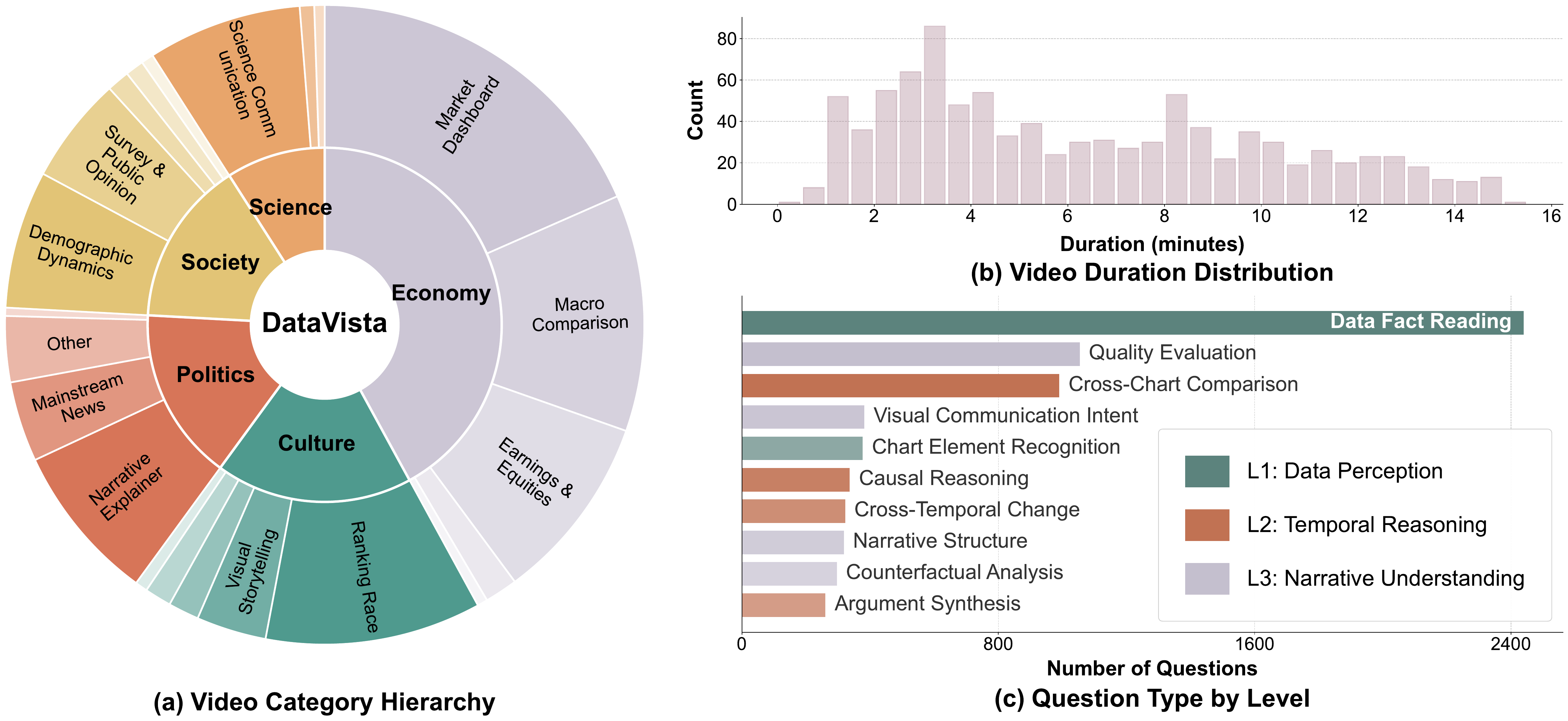}
  \vspace{-2em}
  \caption{Overview of the \benchmark dataset. (a) Video category hierarchy showing the distribution across five major domains and their sub-categories. (b) Distribution of video durations in minutes. (c) Question types organized by evaluation level.}
  \label{fig:dataset_stats}
  \vspace{-1.5em}
\end{figure*}

\paragraph{Topic taxonomy.}
To support fine-grained topic-level evaluation, \rev{we draw on topic classifications in data journalism research~\citep{stalph2018classifying, loosen2020datadriven} to establish a two-level taxonomy with 5 domains (Economy, Society, Science, Politics, and Culture) and 22 subcategories, as shown in Figure~\ref{fig:dataset_stats}a. Videos first receive initial tags based on source channels and search keywords, then are assigned to categories using channel affiliation and title cues. The complete category mapping and subcategory distribution are provided in Appendix~\ref{app:topic_dist}.}

\paragraph{Duration and recency.}
As shown in Figure~\ref{fig:dataset_stats}b, videos range from 28 seconds to 15 minutes, with a median of $\approx$5.6 minutes, matching the typical narrative density reported for professional data videos~\citep{amini2015understanding, DBLP:journals/tvcg/YangXLLGSQC22}. To reduce leakage risk, \benchmark prioritizes recently published content, with approximately 68.5\% of videos published in or after June 2025 (contamination analysis in Appendix~\ref{app:contamination}). For content quality, we use source credibility as the primary gate rather than a single engagement metric, and record view count only as a supplementary indicator. We further manually remove classic films, television episodes, and flagship content of top-tier creators, following the decontamination practice of recent video benchmarks~\citep{fu2025videommev2}.

\paragraph{Screening pipeline.}
\rev{We use Gemini-3-Flash to pre-screen candidate videos (Figure~\ref{fig:benchmark_process}, Stage~1), requiring:
(1)~core information to be data-driven rather than conveyed solely through narration;
(2)~at least one clearly visible and sustained data visualization element;
(3)~a clear information communication intent, excluding pure chart displays, software tutorials, and tool demonstrations.
Annotators then watch the videos to confirm or correct each model classification. The screening prompt, output schema, and verification interface are provided in Appendix~\ref{app:collection}.}
Of the 5{,}604 candidate videos, 967 pass screening and proceed to question construction; after question review, the retained questions cover 961 videos.

\subsection{Capability Hierarchy}
\label{sec:capability}

Understanding a data video requires coordinating multiple abilities, from perceiving data in individual frames, to integrating information across charts over time, to interpreting the creator's narrative intent~\citep{segel2010narrative, amini2015understanding}.
To characterize strengths and bottlenecks in this process, we follow Bloom's Taxonomy~\citep{bloom1956taxonomy}, which is widely used to guide assessment design and orders cognitive objectives from remembering and understanding toward analyzing and evaluating.
We accordingly organize the benchmark questions into three progressive levels comprising 10 fine-grained question types (complete definitions in Appendix~\ref{appendix:capability_hierarchy}).

\paragraph{Level~1: Data Perception.}
This foundational level assesses whether a model can locate relevant visual evidence in a video and read the data it contains, corresponding to the ``reading'' level in visualization literacy research~\citep{DBLP:journals/tvcg/BoyRBF14,DBLP:journals/tvcg/LeeKK17}. Unlike static chart question answering where the target chart is pre-isolated, relevant charts here appear within a continuous video containing multiple visual states, requiring the model to localize evidence before reading; answers do not depend on frame ordering.
It contains two question types: \emph{Data Fact Reading} (reading and cross-frame aggregation of values, trends, and extrema) and \emph{Chart Element Recognition} (chart type, axis, legend, and color encoding).

\paragraph{Level~2: Temporal Reasoning.}
Building on L1, this level assesses the model's ability to integrate evidence across charts and reason about changes and relationships over time~\citep{amini2015understanding,thompson2020understanding,hao2024vistr}.
It contains four question types:
\emph{Cross-Temporal Change}, tracking how events evolve and extrapolate over time;
\emph{Cross-Chart Comparison}, comparing values across charts or deriving, through arithmetic, results not stated explicitly in the video~\citep{cleveland1984graphical};
\emph{Causal Reasoning}, identifying causal chains supported by evidence from multiple charts;
and \emph{Argument Synthesis}, identifying the key evidence that supports the video's central argument.

\paragraph{Level~3: Narrative Understanding.}
\rev{This highest level assesses the model's understanding of how data are organized and presented as a narrative, as well as overall video quality.}
It contains four question types:
\emph{{Narrative Structure}}, covering narrative pattern recognition, data-flow comprehension, and \rev{the roles and connections of non-adjacent segments within the overall argument}~\citep{DBLP:journals/tvcg/YangXLLGSQC22};
\emph{Visual Communication Intent}, analyzing how chart and animation design serves narrative purposes~\citep{shi2021communicating,shen2025reflecting};
\emph{Counterfactual Analysis}, reasoning about how changes in design choices would affect information delivery;
and \emph{Quality Evaluation}, making holistic judgments along the Fidelity--Expressiveness--Aesthetics dimensions~\citep{xie2026visjudgebench}.

\subsection{Question Design}
\label{sec:question_design}

\paragraph{Question format.}
\benchmark contains both four-option single-answer and multiple-select questions.
Given the inherently combinatorial nature of causal argumentation and narrative design in data videos~\citep{DBLP:journals/tvcg/YangXLLGSQC22}, Causal Reasoning, Argument Synthesis, {Narrative Structure}, Visual Communication Intent, and Counterfactual Analysis often require models to identify the correct subset from multiple visual evidence items or design options, and multiple-select questions concentrate in these types.
\rev{Figure~\ref{fig:dataset_stats}c shows the distribution of question types across levels, {with statistics on per-level question counts and the distribution of questions per video in Appendix~\ref{app:question_dist}}.}

\paragraph{Distractor design.}
Each level follows differentiated distractor design principles.
L1 numerical questions include at least one numerically close distractor to test precise reading.
L2 cross-chart questions include at least one distractor that can be selected based on a single chart alone, reducing the shortcut of answering based on a single chart.
L3 narrative questions draw all options from real categories in the narrative design space for data videos~\citep{DBLP:journals/tvcg/YangXLLGSQC22}, avoiding out-of-domain fake options.
Quality evaluation questions follow the Fidelity--Expressiveness--Aesthetics framework~\citep{xie2026visjudgebench}: all options share the same positive tone and differ only in the specific visual mechanism identified, reducing the shortcut of answering based on sentiment polarity.
The complete distractor design principles for each question type are provided in Appendix~\ref{appendix:capability_hierarchy}.

\subsection{Expert Annotation and Quality Control}
\label{sec:annotation}

\paragraph{Candidate question design.}
The annotation workflow is illustrated in Figure~\ref{fig:benchmark_process} (Stages 2--3).
Five annotators with experience in data video production or data visualization research design the candidate questions. To assist this process, Gemini-3-Flash pre-extracts chart segments, key data points, narrative stages, and argument chapters (prompt and output schema in Appendix~\ref{app:pre_extraction}). Annotators watch each video in full, verify and correct the extracted information segment by segment, supplement omissions, and mark timestamps for key data points. They then write question stems, options, and answers following the three-level capability framework and distractor design principles, with detailed specifications in Appendix~\ref{appendix:capability_hierarchy}.
Annotation time per video ranges from approximately 1 to 2 hours, depending on video duration, chart density, and narrative complexity.

\paragraph{Text-only filtering.}
To reduce shortcuts that do not require video evidence~\citep{fu2025videommev2,bian2025you}, we provide Gemini-3-Flash with only question stems and options, filtering out candidates whose predicted answers exactly match the references. Across five representative models from different model families, filtered-out questions are also easier to answer from text alone, with accuracy 22.9--32.2 percentage points higher than on retained questions. Screening details and cross-model analysis are provided in Appendix~\ref{app:text_only}.

\paragraph{Human review and finalization.}
Four reviewers who did not participate in question design review the candidates retained after text-only filtering. \rev{Tasks are assigned by video, with two reviewers separately checking each question's wording, video evidence for the answer, and distractors. Disagreements are discussed against the original video; revised questions are rechecked and those still ambiguous are removed.} Among candidates retained after text-only filtering, 80.1\% pass review, 15.4\% pass after manual revision, and 4.5\% are rejected. The final dataset contains 6{,}775 questions covering 961 videos. \rev{The complete review procedure, level-specific criteria, and statistics are provided in Appendix~\ref{app:question_review}.}

%% file: src/04_experiments.tex
\section{Experiments}
\label{sec:experiment}

\subsection{Experimental Setup}
\label{sec:experimental_setup}

\paragraph{\rev{Models.}}
\rev{We evaluate 19 MLLMs, comprising five closed-source and 14 open-source models. The open-source models span the Qwen3-VL~\citep{bai2025qwen3vl}, Qwen2.5-VL~\citep{bai2025qwen25vl}, InternVL3.5~\citep{internvl35}, and InternVL3~\citep{zhu2025internvl3} families, with Instruct variants used for the Qwen models. Table~\ref{tab:benchmark} lists all model variants. Unless noted otherwise, later diagnostic analyses in this section use five representative models: Gemini-3.1-Pro, GPT-5, Claude-Sonnet-4.6, Qwen3-VL-32B, and InternVL3.5-38B.}

\paragraph{\rev{Input settings.}}
\rev{To compare models under a uniform visual budget, the main evaluation provides all models with question stems, answer options, and 50 frames uniformly sampled from the full video, without audio. We use the same frame set for conditions without and with subtitles, adding subtitle text in the latter. Models provide answers together with supporting reasons. The impact of varying frame budgets and native video inputs is reported in Section~\ref{sec:frame_budget} and Appendix~\ref{app:native_video}, with evaluation details in Appendix~\ref{app:prompts}.}

\paragraph{\rev{Evaluation metric.}}
\rev{We use per-question exact-match accuracy: single-select answers must be correct, and multi-select predictions must exactly match the ground-truth option set. Overall accuracy is computed by pooling responses from the conditions with and without subtitles. For multi-select questions, we also report option-level precision, recall, and F1 to assess partially correct responses (Appendix~\ref{app:multiselect_diagnostics}). Uncertainty estimation is described in Appendix~\ref{app:statistical_reliability}.}

\paragraph{\rev{Human evaluation.}}
\rev{Six evaluators watch complete videos and listen to the original audio. Questions are assigned by video, with two independent answers per question. Those who wrote or finally reviewed a question do not answer it. Scoring follows the same criterion as for the models.}

\subsection{Overall Performance}

\input{table/tab_main_results}

\paragraph{\rev{Human performance.}}
\rev{Human evaluators achieve 91.7\% overall accuracy, with 97.1\%, 89.0\%, and 86.8\% at L1, L2, and L3, respectively. They perform near ceiling on L1 data reading and maintain high accuracy on temporal reasoning and narrative understanding. Exact agreement between paired responses is 88.6\% overall and 95.5\%, 85.7\%, and 81.3\% across the three levels.}

\paragraph{\rev{Overall results.}}
\rev{As shown in Table~\ref{tab:benchmark}, Gemini-3.1-Pro and GPT-5 achieve overall accuracies of 70.0\% and 67.4\%, respectively, with the former still 21.7 percentage points below the human baseline. The strongest open-source model, Qwen3-VL-32B, reaches 57.3\%, trailing Gemini-3.1-Pro by 12.7 points. Its Thinking mode raises overall accuracy to 62.9\%, still below the leading closed-source models. Gains are concentrated in temporal reasoning, without corresponding improvement in narrative understanding. Per-level results and analysis of both modes are provided in Appendix~\ref{app:thinking}. With subtitles, equal weighting across the five domains leaves the ranking of the five representative models unchanged (Appendix~\ref{app:topic_breakdown}).}

\paragraph{\rev{Subtitle gains and baselines.}}
\rev{Adding subtitles raises overall accuracy by a median of 3.1 percentage points across the 19 models, with the largest gain of 5.1 points for Qwen3-VL-32B; Section~\ref{sec:input_evidence} analyzes how subtitles affect the individual capability levels. Random selection and the most-frequent-answer baseline score 20.8\% and 27.5\%, respectively; their settings and per-level results are provided in Appendix~\ref{app:statistical_baselines}.}

\paragraph{\rev{Cross-level differences.}}
\rev{Temporal reasoning is particularly discriminative in these model comparisons. Gemini-3.1-Pro leads Qwen3-VL-32B by 25.3 percentage points on L2, compared with 10.3 points on L1 and 4.2 on L3. On L3 the gap lies mainly between models and humans: the highest accuracy reached by any model is 81.8\% on L1, 15.3 percentage points below humans, but only 59.2\% on L3, where the gap widens to 27.6 points.}

\paragraph{\rev{Native video input.}}
\rev{Given complete videos with their original audio tracks, Gemini-3.1-Pro and Gemini-3-Flash achieve overall accuracies of 73.5\% and 68.9\%, respectively, both below the human score of 91.7\%. On narrative understanding, they score 59.4\% and 55.9\%, respectively, leaving a substantial gap to the human score of 86.8\%. Detailed comparisons and cases are provided in Appendix~\ref{app:native_video}.}

\subsection{\rev{Input Evidence Analysis}}
\label{sec:input_evidence}

\begin{figure}[t!]
    \centering
    \vspace{-1em}
    \includegraphics[width=\textwidth]{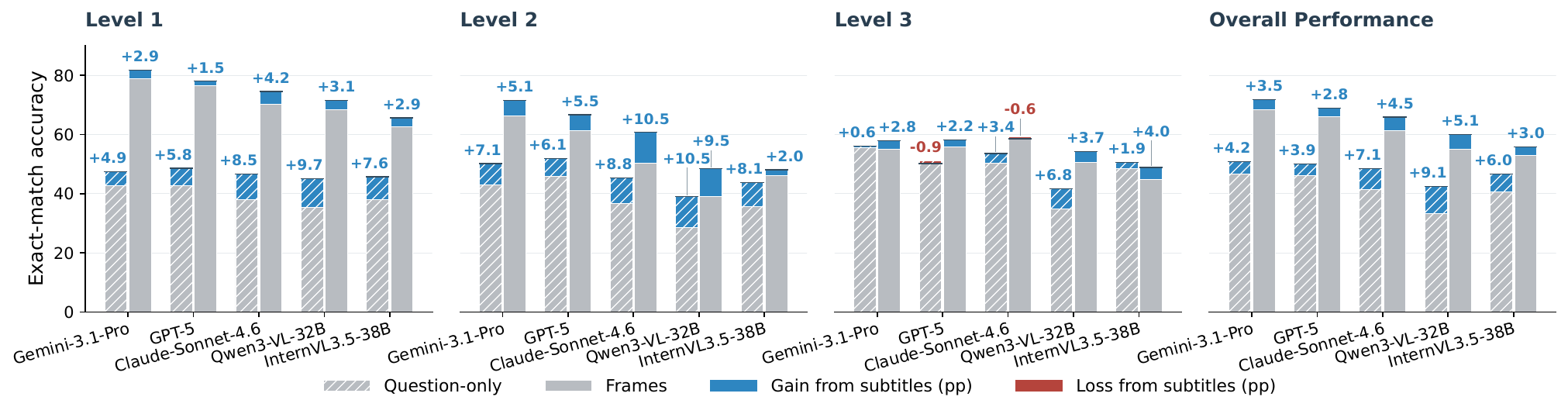}
    \vspace{-2em}
    \caption{\rev{Exact-match accuracy (\%) of five models under question-only and frame inputs, by level. Colored segments show the change after adding subtitles (percentage points).}}
    \label{fig:input_evidence}
    \vspace{-1.5em}
\end{figure}

\rev{To examine the contributions of frames and subtitles, we compare the five representative models under four input conditions: questions only, questions with subtitles, questions with frames, and questions with both frames and subtitles. All conditions include answer options and exclude audio. Figure~\ref{fig:input_evidence} reports the results.}


The contribution of frames decreases across levels. Relative to questions alone, adding frames raises mean L1 accuracy across the five models from 39.4\% to 71.4\%, a gain of 32.0 percentage points, compared with 14.7 on L2 and 5.1 on L3 (from 48.0\% to 53.1\%). On L1, frames alone also outperform questions with subtitles (46.7\%) by 24.7 points, highlighting the importance of visual chart information for basic data reading. These findings indicate that adding frames primarily improves basic chart reading, with relatively limited gains in narrative understanding.

\rev{Subtitles provide the largest supplementary gain for temporal reasoning. Adding subtitles to frames raises mean L2 accuracy from 52.6\% to 59.1\%, a gain of 6.5 percentage points, compared with 2.9 on L1 and 2.4 on L3. Joint input also outperforms questions with subtitles alone by 13.1 points, indicating that frames and subtitles provide complementary information for temporal reasoning.}

\rev{Subtitles can also narrow performance gaps between short and long videos. For GPT-5, adding subtitles raises accuracy on videos longer than 9 minutes from 59.4\% to 66.9\%, reducing the gap from the shortest-video group from 8.9 to 1.4 percentage points. Duration-group results for all models are provided in Appendix~\ref{app:duration_analysis}.}

\subsection{\rev{Frame Budget Sensitivity}}
\label{sec:frame_budget}

\begin{wrapfigure}[13]{r}{0.4\textwidth}
  \vspace{-14pt}
  \centering
  \includegraphics[width=\linewidth]{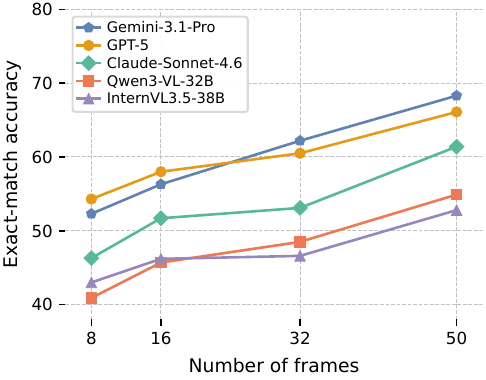}
  \par
  \setlength{\abovecaptionskip}{2pt}
  \caption{\rev{Exact-match accuracy across frame budgets without subtitles or audio.}}
  \label{fig:frame_budget}
\end{wrapfigure}

\rev{Having compared the contributions of frames and subtitles, we further examine the effect of visual sampling density by comparing five representative models at frame budgets of 8, 16, 32, and 50. Frames are sampled at the midpoints of equal temporal segments across the full video, without subtitles or audio.}


\rev{As shown in Figure~\ref{fig:frame_budget}, increasing the frame budget improves overall accuracy and changes which models lead. GPT-5 performs best at 8 and 16 frames, whereas Gemini-3.1-Pro leads at 32 and 50 frames. Among the open-source models, InternVL3.5-38B leads at lower budgets, whereas Qwen3-VL-32B performs better at higher budgets.}

\rev{Basic data perception is more sensitive to sampling density. Increasing the budget from 8 to 32 frames raises mean L1 accuracy across the five models by 11.6 percentage points, compared with 4.7 on L2 and 2.1 on L3. With 100 frames, L1 can still benefit, whereas L2 and L3 show no consistent improvement (Appendix~\ref{app:frame100}). These results indicate that increasing the number of sampled frames primarily improves chart reading, with more limited gains in temporal reasoning and narrative understanding.}

\subsection{Capability Dimension Analysis}

\begin{figure}[t!]
  \centering
  \vspace{-1em}
  \includegraphics[width=\textwidth]{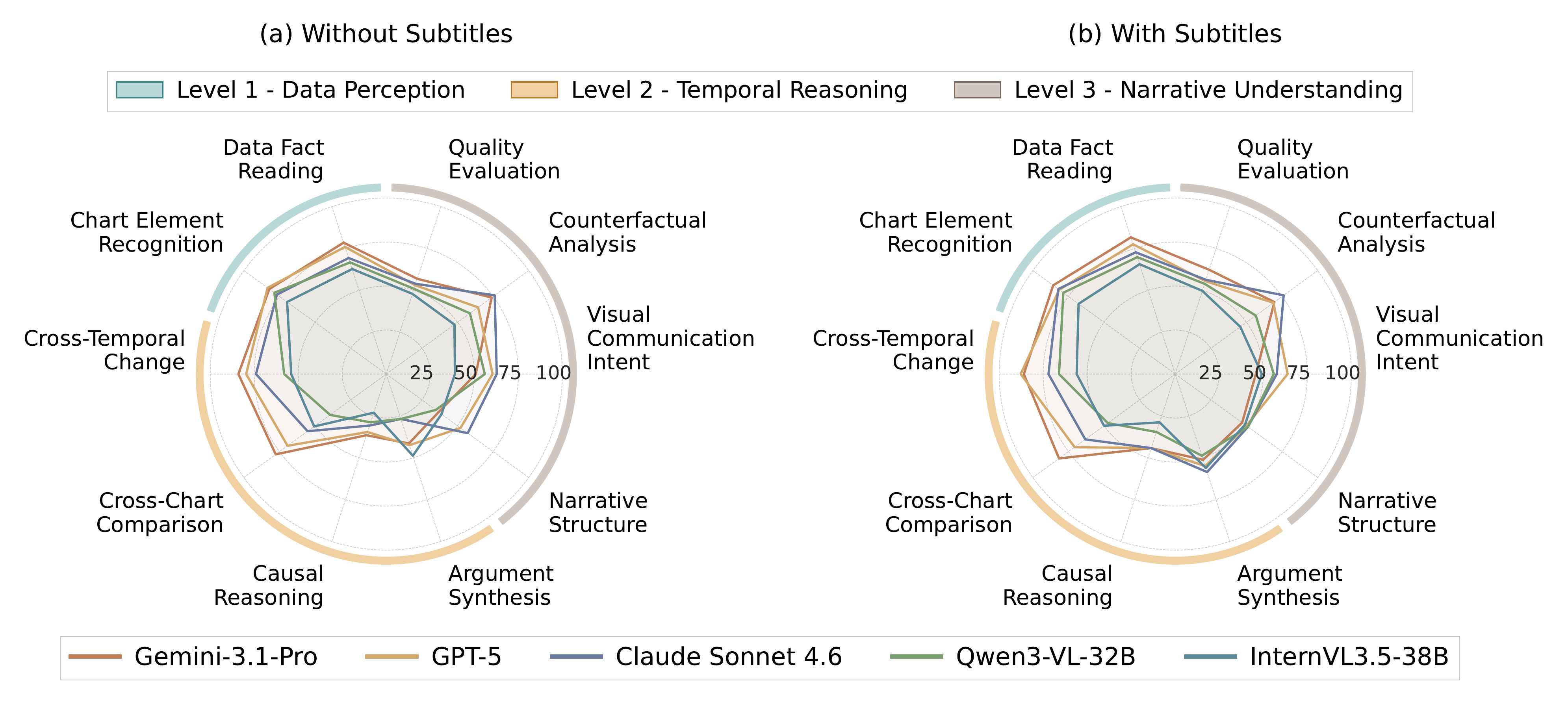}
  \vspace{-2em}
  \caption{\rev{Per-type exact-match accuracy (\%) of five representative models: (a) without subtitles and (b) with subtitles.}}
  \label{fig:capa_radar}
  \vspace{-1.5em}
\end{figure}

We further compare the five representative models across ten question types to examine differences within each capability level (Figure~\ref{fig:capa_radar}).

\paragraph{\rev{Causal and argument tasks.}}
\rev{With both frames and subtitles, mean accuracy is 73.6\% on Cross-Temporal Change and 62.6\% on Cross-Chart Comparison, compared with 39.2\% on Causal Reasoning and 53.9\% on Argument Synthesis; even the best model reaches only 44.2\% on Causal Reasoning. Unlike tracking changes or comparing values, the latter two tasks require identifying which data support a specified causal relationship or the central argument and selecting the complete set of supporting evidence. These two types also benefit most from subtitles, gaining 8.5 and 16.6 percentage points, respectively, compared with at most 4.1 for the other two types.}

\paragraph{\rev{{Narrative structure.}}}
\rev{Across the five models, accuracy on Cross-Temporal Change ranges from 56.0\% to 87.8\%, but all score between 46.9\% and 51.0\% on {Narrative Structure}. These questions further require understanding how data are organized into a narrative and how arguments connect across chapters. Within L3, Counterfactual Analysis averages 63.3\% and {Narrative Structure} averages 49.6\%, indicating that the difficulty of L3 concentrates on {understanding narrative structure}.}

\paragraph{\rev{Answer completeness.}}
For the multi-select questions within these task types, we supplement exact match with option-level F1 to distinguish partially correct from fully correct answers (Appendix~\ref{app:multiselect_diagnostics}). With subtitles, Argument Synthesis reaches 88.0\% F1 but 53.9\% exact match, indicating that many responses are only partially correct and that fully correct answers remain challenging. Even with partial credit, Causal Reasoning has the lowest F1 among the five multi-select types.

\subsection{\rev{Error Analysis}}
\label{sec:error_analysis}

\rev{Following diagnostic frameworks from chart question answering and video understanding~\citep{masry2022chartqa,zhu2025multichartqa,fu2024videomme}, we manually inspect 120 randomly sampled incorrect responses from Gemini-3.1-Pro under 50-frame input with subtitles and group them into perception, reasoning, and instruction understanding.}

\paragraph{\rev{Perceptual errors (47.5\%).}}
The most common cases are value and text reading errors (36.7\%), such as reading the Nintendo Wii first-three-year sales total as 45 million rather than about 20 million. The remaining cases include confused visual mappings (5.0\%), reversed trends or rankings (3.3\%), and chart-structure errors such as misjudged axis scales (2.5\%). These errors occur with the relevant frames already available, indicating that accurately extracting chart information remains a bottleneck in data video understanding.

\paragraph{\rev{Reasoning errors (42.5\%).}}
\rev{After chart facts have been extracted, the main difficulty lies in building logical relations. Pure arithmetic errors are rare (0.8\%); more common are cross-chart linking (4.2\%) and evidence judgment (37.5\%). On causal reasoning and argument synthesis in particular, the model often reads the chart facts correctly but fails to judge whether a given fact supports a specified claim, indicating that it still struggles to connect observed data with the narrative argument.}

\paragraph{\rev{Instruction understanding errors (10.0\%).}}
\rev{These errors do not arise from failing to read the chart, but from not following constraints in the question: the model shifts the specified time range or target object (5.0\%), maps an otherwise reasonable conclusion to the wrong option (3.3\%), or overlooks a restriction on the evidence source (1.7\%), for instance selecting a narrated revenue target when the question asks for chart-specific facts.}

\rev{Further analysis of Qwen3-VL-32B also identifies cases where the model correctly reads chart facts but misjudges whether they support the claim in the question; detailed diagnostics are provided in Appendix~\ref{app:qwen_error_analysis}. Error cases for Gemini-3.1-Pro and example questions across capability levels are provided in Appendix~\ref{app:gemini_error_cases} and Appendix~\ref{appendix:case_studies}, respectively.}

%% file: table/tab_main_results.tex
\begin{table*}[t!]
\centering
\vspace{-1em}
\caption{\rev{Exact-match accuracy (\%) on \benchmark. \emph{w/o Sub.} and \emph{w/ Sub.} use the same 50 sampled frames without audio, without and with subtitles; \emph{Avg Acc} weights the three levels by question count, and \emph{Overall} pools both conditions. Human evaluation uses complete videos with original audio.}}
\label{tab:benchmark}
\begingroup
\renewcommand{\cmark}[2]{%
  \ifdim #1pt > 60pt \cellcolor{heatA!45}#2%
  \else\ifdim #1pt > 50pt \cellcolor{heatA!30}#2%
  \else\ifdim #1pt > 40pt \cellcolor{heatA!18}#2%
  \else\ifdim #1pt > 30pt \cellcolor{heatA!8}#2%
  \else \ifshowrevisions #2\else{\color{black!50}#2}\fi%
  \fi\fi\fi\fi
}
\ifshowrevisions\color{blue}\fi
\renewcommand{\arraystretch}{1.15}
\setlength{\tabcolsep}{4pt}
\resizebox{\textwidth}{!}{%
\begin{tabular}{@{}l cc cc cc cc !{\color{gray}\vrule width 0.6pt} c@{}}
\toprule
 & \multicolumn{2}{c}{Level 1} & \multicolumn{2}{c}{Level 2} & \multicolumn{2}{c}{Level 3} & \multicolumn{2}{c!{\color{gray}\vrule width 0.6pt}}{Avg Acc} & \multirow{2}{*}{\textbf{Overall}} \\
\cmidrule(lr){2-3}\cmidrule(lr){4-5}\cmidrule(lr){6-7}\cmidrule(lr){8-9}
Model & w/o Sub. & w/ Sub. & w/o Sub. & w/ Sub. & w/o Sub. & w/ Sub. & w/o Sub. & w/ Sub. & \\
\midrule
\multicolumn{10}{c}{\textit{Human Baseline}} \\
\midrule
\rowcolor{humanrow}
Human Expert & \multicolumn{2}{c}{97.1} & \multicolumn{2}{c}{89.0} & \multicolumn{2}{c}{86.8} & \multicolumn{2}{c}{91.7} & -- \\
\midrule
\multicolumn{10}{c}{\textit{Closed-Source Models}} \\
\midrule
Gemini-3.1-Pro & \textbf{78.9} & \textbf{81.8} & \textbf{66.4} & \textbf{71.5} & 55.1 & 57.9 & \textbf{68.2} & \textbf{71.7} & \cmark{70.0}{\textbf{70.0}} \\
GPT-5 & 76.5 & 78.0 & 61.2 & 66.7 & 55.9 & 58.1 & 66.0 & 68.8 & \cmark{67.4}{67.4} \\
Gemini-3-Flash & 75.6 & 78.0 & 59.7 & 68.8 & 52.3 & 55.1 & 64.1 & 68.5 & \cmark{66.3}{66.3} \\
GPT-4o & 65.0 & 66.4 & 42.5 & 44.8 & 56.6 & \textbf{58.7} & 56.1 & 58.0 & \cmark{57.1}{57.1} \\
Claude-Sonnet-4.6 & 70.3 & 74.5 & 50.2 & 60.7 & \textbf{59.2} & 58.6 & 61.3 & 65.8 & \cmark{63.6}{63.6} \\
\midrule
\multicolumn{10}{c}{\textit{Open-Source Models}} \\
\midrule
Qwen3-VL-32B-Instruct & 68.5 & 71.6 & 39.0 & 48.5 & 50.5 & 54.2 & 54.7 & 59.8 & \cmark{57.3}{57.3} \\
InternVL3.5-38B & 62.7 & 65.6 & 46.1 & 48.1 & 44.9 & 48.9 & 52.6 & 55.6 & \cmark{54.1}{54.1} \\
InternVL3.5-14B & 54.5 & 60.3 & 36.9 & 42.7 & 40.2 & 41.4 & 45.2 & 49.6 & \cmark{47.4}{47.4} \\
InternVL3.5-8B & 54.5 & 60.5 & 35.9 & 36.3 & 39.3 & 39.6 & 44.7 & 47.4 & \cmark{46.1}{46.1} \\
Qwen3-VL-8B-Instruct & 59.0 & 64.7 & 28.5 & 33.9 & 38.0 & 38.9 & 44.1 & 48.2 & \cmark{46.2}{46.2} \\
InternVL3-8B & 55.2 & 59.0 & 33.7 & 38.2 & 36.2 & 37.7 & 43.4 & 46.7 & \cmark{45.1}{45.1} \\
Qwen2.5-VL-32B-Instruct & 63.4 & 65.4 & 46.4 & 47.8 & 48.0 & 48.9 & 54.0 & 55.5 & \cmark{54.8}{54.8} \\
Qwen2.5-VL-7B-Instruct & 56.3 & 63.2 & 33.6 & 33.9 & 31.2 & 31.8 & 42.3 & 45.4 & \cmark{43.9}{43.9} \\
InternVL3.5-4B & 53.7 & 48.3 & 35.9 & 33.6 & 39.9 & 35.8 & 44.5 & 40.4 & \cmark{42.5}{42.5} \\
Qwen3-VL-4B-Instruct & 57.9 & 61.0 & 25.4 & 30.5 & 34.9 & 37.7 & 41.8 & 45.4 & \cmark{43.6}{43.6} \\
Qwen2.5-VL-3B-Instruct & 47.7 & 53.0 & 35.3 & 36.9 & 27.1 & 27.7 & 38.0 & 40.8 & \cmark{39.4}{39.4} \\
InternVL3.5-2B & 47.7 & 52.1 & 27.8 & 29.8 & 27.1 & 28.0 & 35.9 & 38.5 & \cmark{37.2}{37.2} \\
Qwen3-VL-2B-Instruct & 50.3 & 57.4 & 28.8 & 30.5 & 25.2 & 24.9 & 36.7 & 40.0 & \cmark{38.4}{38.4} \\
InternVL3-2B & 40.5 & 44.6 & 25.0 & 21.8 & 23.5 & 21.1 & 31.0 & 31.1 & \cmark{31.1}{31.1} \\
\bottomrule
\end{tabular}%
}
\endgroup
\vspace{-1.5em}
\end{table*}

%% file: src/05_conclusion.tex
\section{Conclusion}
\label{sec:conclusion}

\rev{We present \benchmark, the first benchmark for data video understanding, comprising 961 real-world data videos and 6{,}775 questions across three capability levels and 10 question types. Evaluation of 19 MLLMs reveals a substantial gap from human performance, with persistent difficulties in identifying evidence for causal claims and understanding narratives. \benchmark provides a foundation for systematically evaluating and improving data video understanding.}

%% file: src/ai_use_statement.tex
\subsection*{AI use statement}
\rev{In this work, during dataset construction we used Gemini-3-Flash for data-video pre-screening, structured information pre-extraction, and text-only filtering based on question stems and options alone. These outputs were verified or corrected by humans, and questions were written by annotators. We used GPT-6 Astra for English grammar polishing and for suggestions on wording. We did not use generative AI to create synthetic data, propose theoretical models or mathematical claims, write proofs, or introduce novel algorithmic ideas or academic claims. The remaining required-disclosure tasks are not applicable to this work. All AI-assisted content was reviewed by the authors, who take full responsibility for the final content. Prompts contained no private or sensitive data. Large language models are not authors of this paper.}

%% file: src/ethics_statement.tex
\subsection*{Ethics statement}
\rev{This work introduces DataVista, an academic benchmark for data video understanding. The source videos used to construct this benchmark were collected from public video platforms (e.g., YouTube). To advance research on multimodal large language models in this emerging domain, we open-source all high-quality annotations meticulously created by our team, including question-answer pairs, evidence timestamps, and structured metadata. The copyright of all source videos belongs to their respective original creators or rightsholders, and this benchmark along with its open-source data is intended solely for academic research. Furthermore, during the dataset construction process, we ensured that all participating domain experts received fair compensation. The data collection and annotation procedures of this study have been reviewed and approved by our institutional ethics board. We hope DataVista will serve as a valuable evaluation tool to promote the development of automated data visualization assessment and understanding technologies.}

%% file: src/reproducibility_statement.tex
\subsection*{Reproducibility statement}
\rev{To support reproducibility, we release the question annotations and video source records. The project repository is available at \url{https://github.com/HKUSTDial/DataVista}. Detailed descriptions of our methodology are provided throughout the paper and appendices: video collection, screening, and question design are detailed in Section~\ref{sec:construction} and Appendix~\ref{appendix:dataset_stats}; the three-level capability framework is defined in Appendix~\ref{appendix:capability_hierarchy}; {evaluation prompts and input settings (including frame sampling strategies) are provided in Appendix~\ref{app:prompts}, and scoring rules in Section~\ref{sec:experimental_setup}}; and data contamination analysis is discussed in {Appendix~\ref{app:contamination}}. To ensure long-term consistency and standardization of future model evaluations, we will deploy an official automated evaluation service and a public leaderboard upon publication, allowing researchers to seamlessly benchmark their models and track ongoing progress in data video understanding.}

%% file: src/appendix-a-Capability_Hierarchy.tex
\section{Capability Hierarchy and Question Type Specification}
\label{appendix:capability_hierarchy}

This appendix provides the complete definitions of all 10 question types in \benchmark, organized by capability level. Each question type is specified along four aspects: (i)~definition, (ii)~target competency, (iii)~required visual evidence, and (iv)~distractor design principles. These specifications served as annotation guidelines during dataset construction and as the reference framework for the fine-grained capability analysis presented in Section~\ref{sec:experiment}.

\begin{figure*}[t!]
  \centering
  \includegraphics[width=\textwidth]{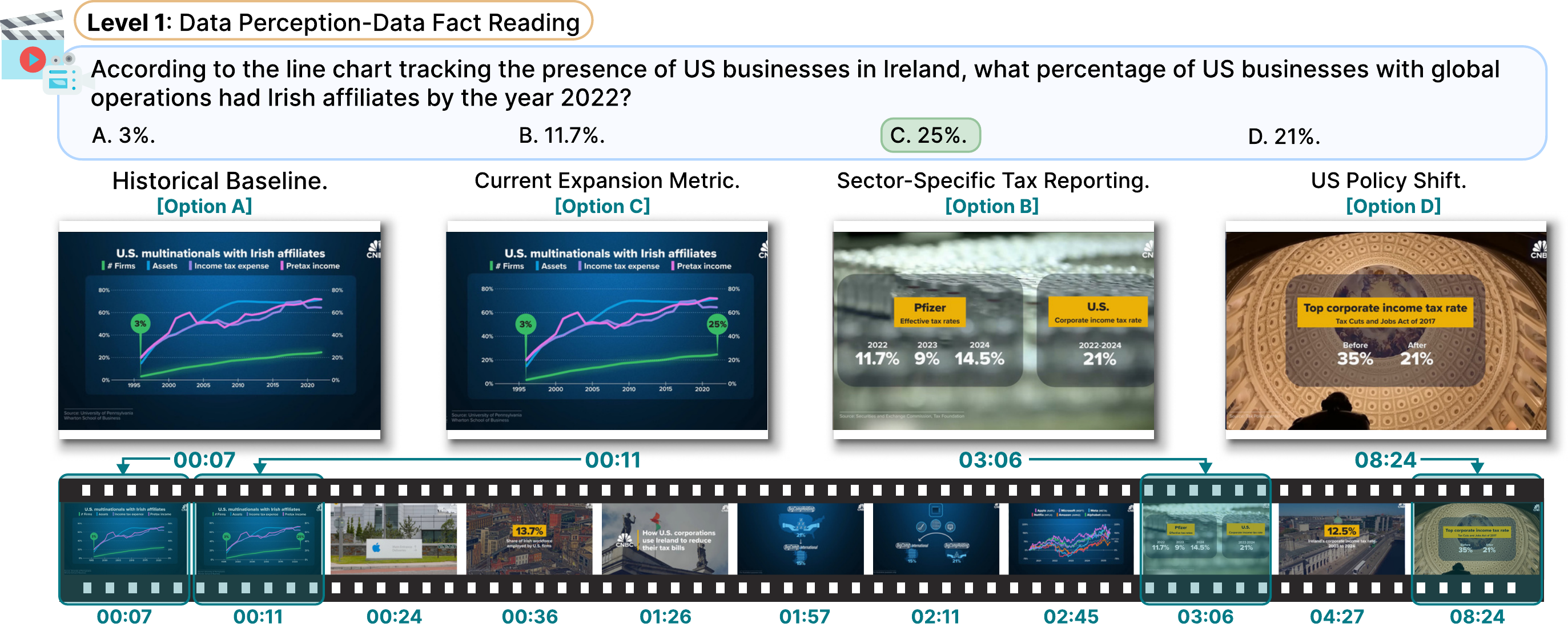}
  \caption{An example L1 question in \benchmark (Data Fact Reading). The question requires extracting a specific numerical value from a trend chart. The correct answer (C) is directly grounded in the visual label for the year 2022, while distractors correspond to the historical baseline (A), \rev{sector-specific tax reporting (B), and the US policy shift (D)}.}
  \label{fig:dvbench_l1_example}
\end{figure*}

\begin{figure*}[t]
  \centering
  \includegraphics[width=\textwidth]{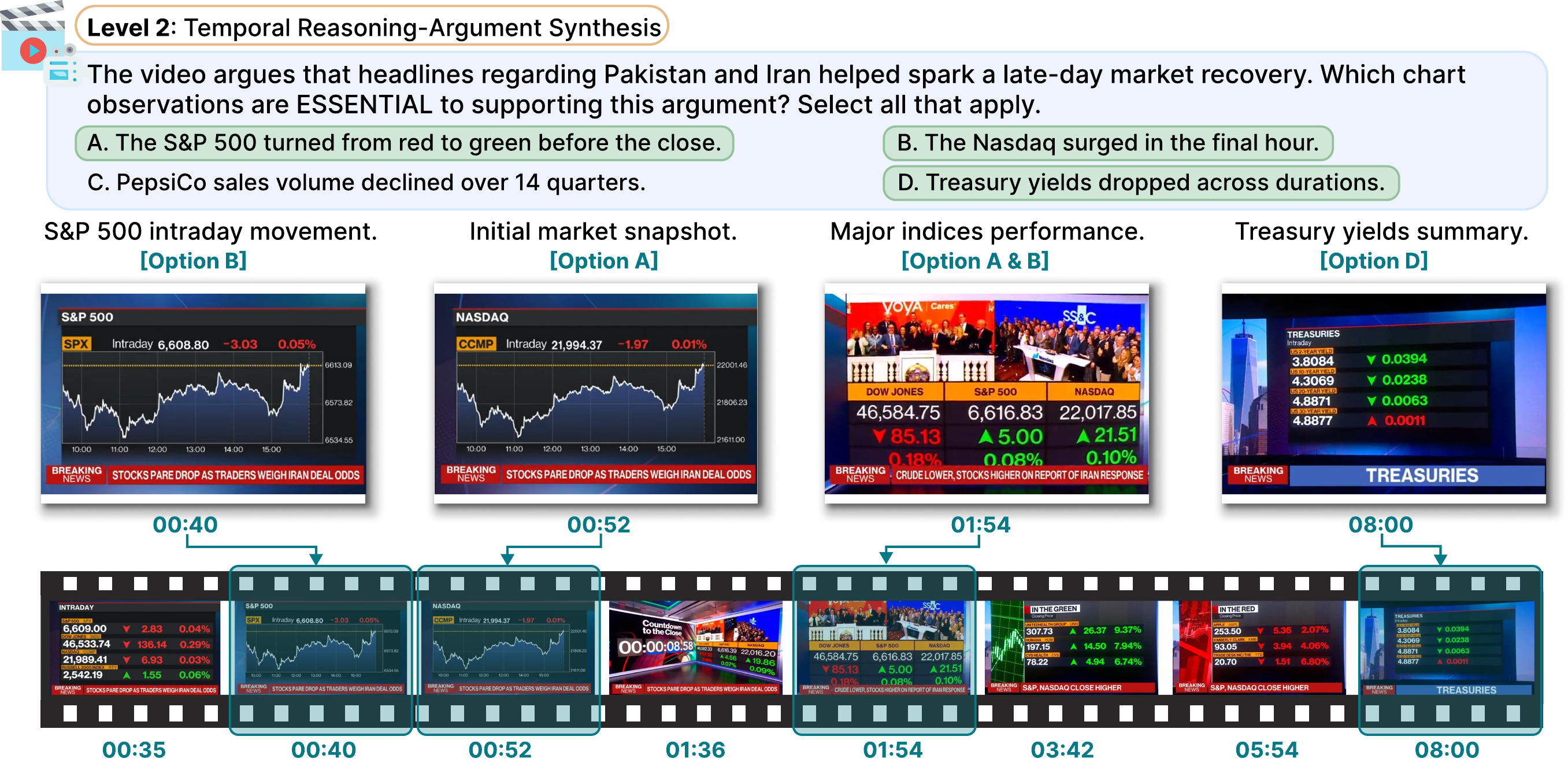}
  \caption{An example L2 question in \benchmark. The video argues that headlines regarding Pakistan and Iran helped spark a late-day market recovery. The task is to select all essential chart observations (A, B, D) that support the argument.}
  \label{fig:dvbench_l2_example}
\end{figure*}

\begin{figure*}[t]
  \centering
  \includegraphics[width=\textwidth]{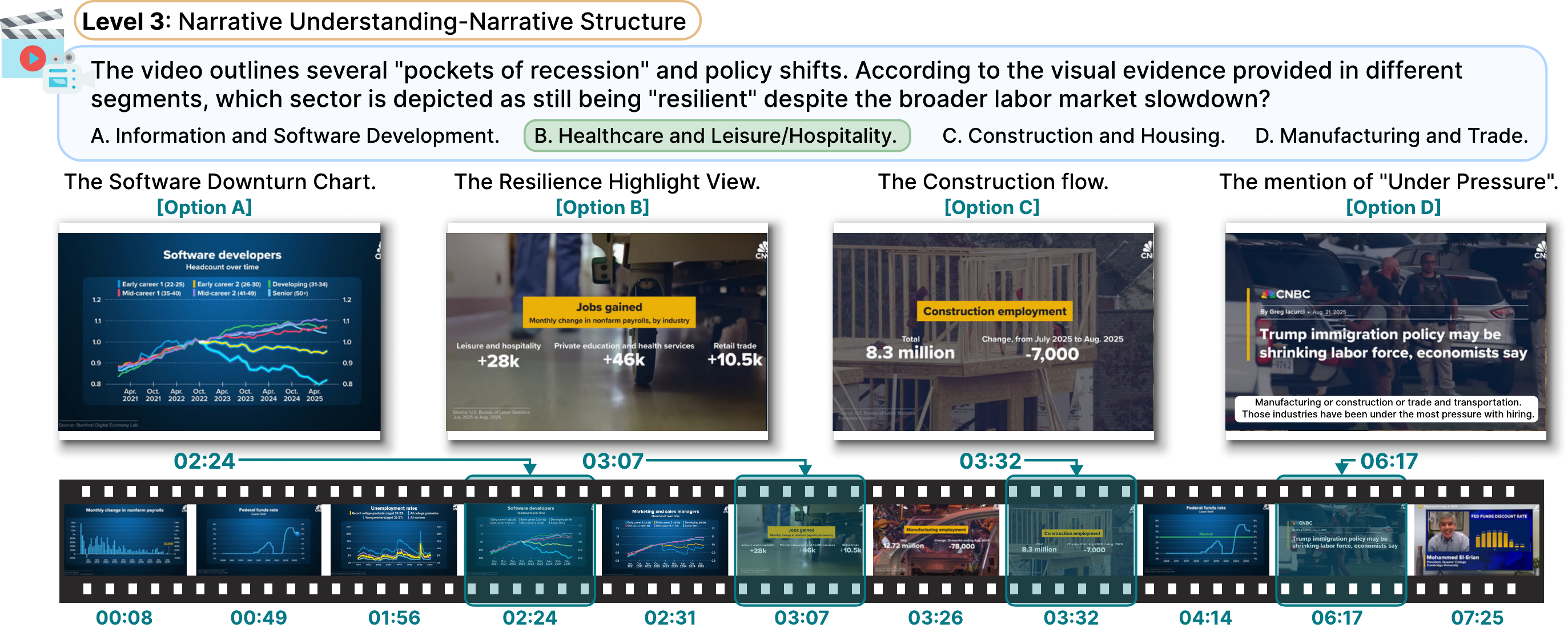}
  \caption{An example L3 question in \benchmark ({Narrative Structure}). This task evaluates the model's ability to synthesize global narrative themes. \rev{The model must distinguish the sector portrayed as still resilient (B: healthcare and leisure/hospitality) from the other narrative roles: the software-employment downturn (A), the decline in construction employment (C), and manufacturing, construction, and trade, which the video explicitly describes as under pressure (D).}}
  \label{fig:dvbench_l3_example}
\end{figure*}

\subsection{Design Rationale}
\label{appendix:design_rationale}

\rev{The three progressive levels and ten question types in \benchmark are designed around capability organization and evidence requirements.}
 
\paragraph{\rev{Capability organization.}}
\rev{The three levels are distinguished by the judgments required: L1 focuses on reading data facts and chart elements; L2 on data relationships and evidential support across charts and time points; and L3 on how this information is organized into a narrative and how visual design serves communication. This design draws on the visualization literacy frameworks proposed by \citet{DBLP:journals/tvcg/BoyRBF14} and \citet{DBLP:journals/tvcg/LeeKK17}, extending them to cross-chart relationships and narrative context in data videos. For example, Argument Synthesis focuses on identifying key data supporting the central claim, whereas {Narrative Structure} focuses on the roles and organization of different segments within the overall argument.}
 
\paragraph{Evidence requirements.}
\rev{Evidence requirements depend on the specific task. Value retrieval typically relies on the relevant chart state~\citep{yang2024askchart}, cross-chart comparison requires combining data from different charts, and narrative and design judgments also require the corresponding context~\citep{wang2026tabletale}. Questions about animation or presentation order require examining how the visual content changes over time. Multiple frames support aggregation or comparison, while order information supports judgments about change processes and narrative organization.}

\rev{Section~\ref{appendix:leaf_spec} specifies each question type's definition, target competency, evidence requirements, and distractor design. Individual types can include multiple modes: Cross-Chart Comparison covers visual comparison and arithmetic derivation; {Narrative Structure} covers narrative patterns, data flow, and cross-section argumentation.}

\subsection{Question Type Specification}
\label{appendix:leaf_spec}

The following gives a complete specification of all 10 question types. \rev{The \textit{Evidence} field describes the visual information and context targeted by each question type, including relevant chart states, cross-chart evidence, and temporal or narrative relationships.} Among these, Data Fact Reading, Cross-Temporal Change, Cross-Chart Comparison, and {Narrative Structure} each include multiple assessment modes, detailed in their respective definitions.

\begin{examplebox}{Level 1: Data Perception}
\label{level1}

\textbf{Data Fact Reading}
    \begin{itemize}
        \item \textit{Definition:} Reading data information from charts, including four modes: (i)~value retrieval, extracting specific values annotated in the chart; (ii)~trend identification, determining the direction and shape of variable changes; (iii)~extremum detection, locating maxima, minima, peaks, or valleys; (iv)~statistical aggregation, aggregating information across multiple frames (e.g., counting the amount of distinct chart types appearing in a video), independent of frame order.
        \item \textit{Target Competency:} \rev{Reading chart data and identifying visual variables; aggregating information from multiple video frames.}
        \item \textit{Evidence:} Single frame / multi-frame unordered (aggregation mode).
        \item \textit{Distractor Design:} At least one distractor has a value within $\pm15\%$ of the correct answer; at least one distractor is drawn from another visible data point in the same chart; aggregation-mode options are consecutive integers ($N{-}1, N, N{+}1, N{+}2$).
    \end{itemize}

    \textbf{Chart Element Recognition}
    \begin{itemize}
        \item \textit{Definition:} Recognizing basic visual elements of a chart, including chart type, axis labels, units, legends, color encodings, etc.
        \item \textit{Target Competency:} Chart type identification; visual encoding comprehension.
        \item \textit{Evidence:} Single frame.
        \item \textit{Distractor Design:} For chart-type questions, at least one distractor is a visually similar chart family member~\citep{wu2025chartretrieval,wang2025vizta} (e.g., area chart vs.\ line chart, stacked bar vs.\ grouped bar); for axis-label questions, at least one distractor is a label from another chart in the same video.
    \end{itemize}

\end{examplebox}

\noindent\textbf{Example.} Figure~\ref{fig:dvbench_l1_example} shows a representative L1 question (Data Fact Reading, value retrieval mode). The question asks for a specific percentage from a trend chart; the correct answer is directly grounded in a visual label at a single frame, while distractors are drawn from numerically close values in the same chart or related metrics from adjacent frames. This example illustrates how L1 questions test precise data reading without requiring temporal reasoning or cross-chart integration.

\begin{examplebox}{Level 2: Temporal Reasoning}

\textbf{Cross-Temporal Change}
    \begin{itemize}
        \item \textit{Definition:} Cross-temporal data change reasoning, including two modes: (i)~change tracking, recording the direction, magnitude, or timing of data changes within the visible time window; (ii)~trend extrapolation, predicting future values beyond the video's time range based on observed trends.
        \item \textit{Target Competency:} Temporal change tracking; trend extrapolation.
        \item \textit{Evidence:} \rev{Data values or chart states at relevant time points, together with their temporal relationships.}
        \item \textit{Distractor Design:} At least one distractor describes a value from only one time point rather than a temporal difference; at least one distractor gives the opposite direction of change; all options in extrapolation mode must be specific numerical values or narrow ranges, not qualitative descriptions.
    \end{itemize}

\textbf{Cross-Chart Comparison}
    \begin{itemize}
        \item \textit{Definition:} Comparing data relationships across charts, including two modes: (i)~visual comparison, determining relative relationships among data in two or more charts by visual inspection alone; (ii)~arithmetic derivation, performing arithmetic operations (ratios, products, proportions, etc.) on values from multiple sources to obtain derived values not explicitly stated in the video.
        \item \textit{Target Competency:} Cross-chart integration; cross-source numerical derivation.
        \item \textit{Evidence:} \rev{Relevant values and visual evidence from at least two chart segments.}
        \item \textit{Distractor Design:} At least one distractor is correct for one chart but incorrect when combined with another (single-chart trap); arithmetic-derivation distractors correspond to distinct computational errors (wrong input value, reversed operation, using complement, etc.); all options must agree in thematic direction, differing only in precise visual details.
    \end{itemize}

\textbf{Causal Reasoning}
    \begin{itemize}
        \item \textit{Definition:} Identifying causal chains jointly supported by multiple charts in the video. \rev{Each option is an independently verifiable visual observation.}
        \item \textit{Target Competency:} Cross-chart causal evidence identification.
        \item \textit{Evidence:} \rev{Observations from at least two chart segments that support the stated causal relationship, including temporal relationships relevant to the claim.}
        \item \textit{Distractor Design:} Correct options must span $\geq$2 chart segments; distractors fall into two types: (A)~observations not shown in the video but causally plausible, and (B)~details genuinely present in the video but not directly relevant to the stated causal relationship.
    \end{itemize}

 \textbf{Argument Synthesis}
    \begin{itemize}
        \item \textit{Definition:} Identifying the key evidence in the video's core argument. \rev{Each option is a data fact from a specific chart segment.}
        \item \textit{Target Competency:} Full video argument synthesis; key evidence identification.
        \item \textit{Evidence:} \rev{Chart evidence across multiple narrative stages, considered in the context of the video's central argument.}
        \item \textit{Distractor Design:} Correct options must span $\geq$2 narrative stages; distractors fall into two types: (A)~facts genuinely present in the video but serving only as supporting details rather than key evidence, and (B)~evidence that contradicts the argument's direction or belongs to a different claim.
    \end{itemize}

\end{examplebox}

\noindent\textbf{Example.} Figure~\ref{fig:dvbench_l2_example} shows a representative L2 question (Argument Synthesis). The video argues that geopolitical headlines sparked a late-day market recovery; the model must select all essential chart observations (from multiple chart segments at different timestamps) that jointly support this argument. Correct options span at least two distinct chart segments, while distractors include facts genuinely present in the video but irrelevant to the stated causal claim. \rev{This example shows that Argument Synthesis requires integrating chart evidence from multiple charts at different temporal positions, going beyond single-frame perception.}

\begin{examplebox}{Level 3: {Narrative Understanding}}

\textbf{{Narrative Structure}}
    \begin{itemize}
        \item \textit{Definition:} Understanding the video's {narrative structure}, including three modes: (i)~pattern recognition, identifying narrative patterns from the design space of \citet{DBLP:journals/tvcg/YangXLLGSQC22}; (ii)~data flow comprehension, understanding the organizational order and presentation strategy of data facts; (iii)~cross section argumentation, integrating data evidence from two or more non-adjacent argumentative segments to understand the video's overall argument structure.
        \item \textit{Target Competency:} Narrative structure comprehension; cross section synthesis.
        \item \textit{Evidence:} \rev{Multiple narrative stages or argument sections, including their presentation sequence for narrative organization and their evidence relationships for cross-section argumentation.}
        \item \textit{Distractor Design:} For pattern recognition and data flow modes, all options must be genuine categories from the same stage of the data stories design space~\citep{DBLP:journals/tvcg/YangXLLGSQC22}\rev{, which organizes 14 narrative patterns and their associated data flows across the Setting, Rising-Climax, and Resolution stages (for example, Statistic Hook in the Setting stage, Showing Contrast in the Rising-Climax stage, and Recap in the Resolution stage)}; cross section distractors are over-generalizations of a single section's claim or misattributions of argumentative structure.
    \end{itemize}

\textbf{Visual Communication Intent}
    \begin{itemize}
        \item \textit{Definition:} Understanding how chart design choices and animation strategies serve data narrative intent.
        \item \textit{Target Competency:} Design intent comprehension; visual mechanism attribution.
        \item \textit{Evidence:} \rev{Relevant chart-design features and their narrative context, including the animation sequence for questions about animation strategies.}
        \item \textit{Distractor Design:} \rev{Options are drawn from the visual communication categories of the same design space; for questions about animation strategies, options are drawn from the six categories of animation techniques, namely entrance, emphasis, exit, transition, data-driven, and temporal animation~\citep{shi2021communicating}.} At least one distractor correctly identifies the visual mechanism but attributes it to the wrong narrative purpose (correct mechanism, wrong intent); at least one distractor correctly identifies the narrative purpose but attributes it to the wrong visual mechanism (correct intent, wrong mechanism).
    \end{itemize}

\textbf{Counterfactual Analysis}
    \begin{itemize}
        \item \textit{Definition:} Reasoning about the impact on communication goals when a specific design choice is changed.
        \item \textit{Target Competency:} Design counterfactual reasoning.
        \item \textit{Evidence:} \rev{The original design and its communication context, including temporal sequences when the proposed change concerns animation or narrative order.}
        \item \textit{Distractor Design:} Alternative designs must be genuinely feasible, not ``straw man'' proposals; extreme options such as ``completely incomprehensible'' or ``no difference at all'' are prohibited; \rev{the number of correct answers is not fixed} to prevent models from exploiting a fixed answer count.
    \end{itemize}

\textbf{Quality Evaluation}
    \begin{itemize}
        \item \textit{Definition:} Making holistic quality assessments of the video's data presentation, with three variants: the Fidelity variant evaluates data presentation faithfulness; the Expressiveness variant evaluates information communication effectiveness; the Aesthetics variant evaluates visual design quality.
        \item \textit{Target Competency:} Holistic assessment of fidelity, expressiveness, and aesthetics.
        \item \textit{Evidence:} \rev{Chart content and visual presentation across the relevant video segments; full-video context for the Expressiveness variant.}
        \item \textit{Distractor Design:} All options share the same positive tone, differing only in the specific visual mechanisms identified; extreme wordings such as ``severe distortion,'' ``complete failure,'' or ``flawless'' are prohibited; distractors must point to visual elements genuinely present in the video, not fabricated issues. \rev{When a question targets potentially misleading design, the mechanisms considered are drawn from visually detectable misleader categories such as truncated axes, unconventional scale directions, and cherry-picked time ranges~\citep{ge2023calvi}, and are used only when the issue is genuinely visible in the video.}
    \end{itemize}

\end{examplebox}

\noindent\textbf{Example.} Figure~\ref{fig:dvbench_l3_example} shows a representative L3 question ({Narrative Structure}, cross-section argumentation mode). The video discusses ``pockets of recession'' across multiple labor-market segments presented in non-adjacent sections; the model must synthesize information from these dispersed segments to identify which sector the video characterizes as ``resilient.'' Distractors correspond to sectors that are genuinely discussed but serve a different narrative role (sectors under pressure rather than resilient ones). This example highlights how L3 questions demand understanding the creator's narrative design and integrating thematic roles across the full video structure.

%% file: src/appendix-b-dataset_statistics.tex
\section{\rev{Dataset Construction and Statistics}}
\label{appendix:dataset_stats}

\subsection{Video Collection Pipeline}
\label{app:collection}

We collect videos from YouTube through two complementary strategies: channel-level crawling and keyword search. Channel-level crawling covers professional media (Bloomberg, Financial Times, The Economist, WSJ), data journalism organizations (Vox, Our World in Data, Pew Research Center), educational channels (Kurzgesagt, TED-Ed, Khan Academy, 3Blue1Brown), and independent data visualization creators. Keyword search uses multiple query groups (e.g., ``data-driven story explanation chart,'' ``climate change graph explained,'' ``election results map breakdown,'' ``public health statistics video'') spanning all five topic domains to supplement long-tail content that channel crawling cannot easily reach.

Candidate videos undergo automated filtering (duration range, upload date threshold) and deduplication, followed by round-robin sampling by provisional topic tags. Each video is registered with structured metadata upon ingestion, including upload date, view count, channel information, YouTube category tags, duration, and available subtitle languages, supporting the contamination analysis (Appendix~\ref{app:contamination}) and statistical analyses below.

\paragraph{AI screening and human verification.}

\begin{figure}[!htbp]
    \centering \includegraphics[width=0.75\linewidth,height=0.85\textheight,keepaspectratio]{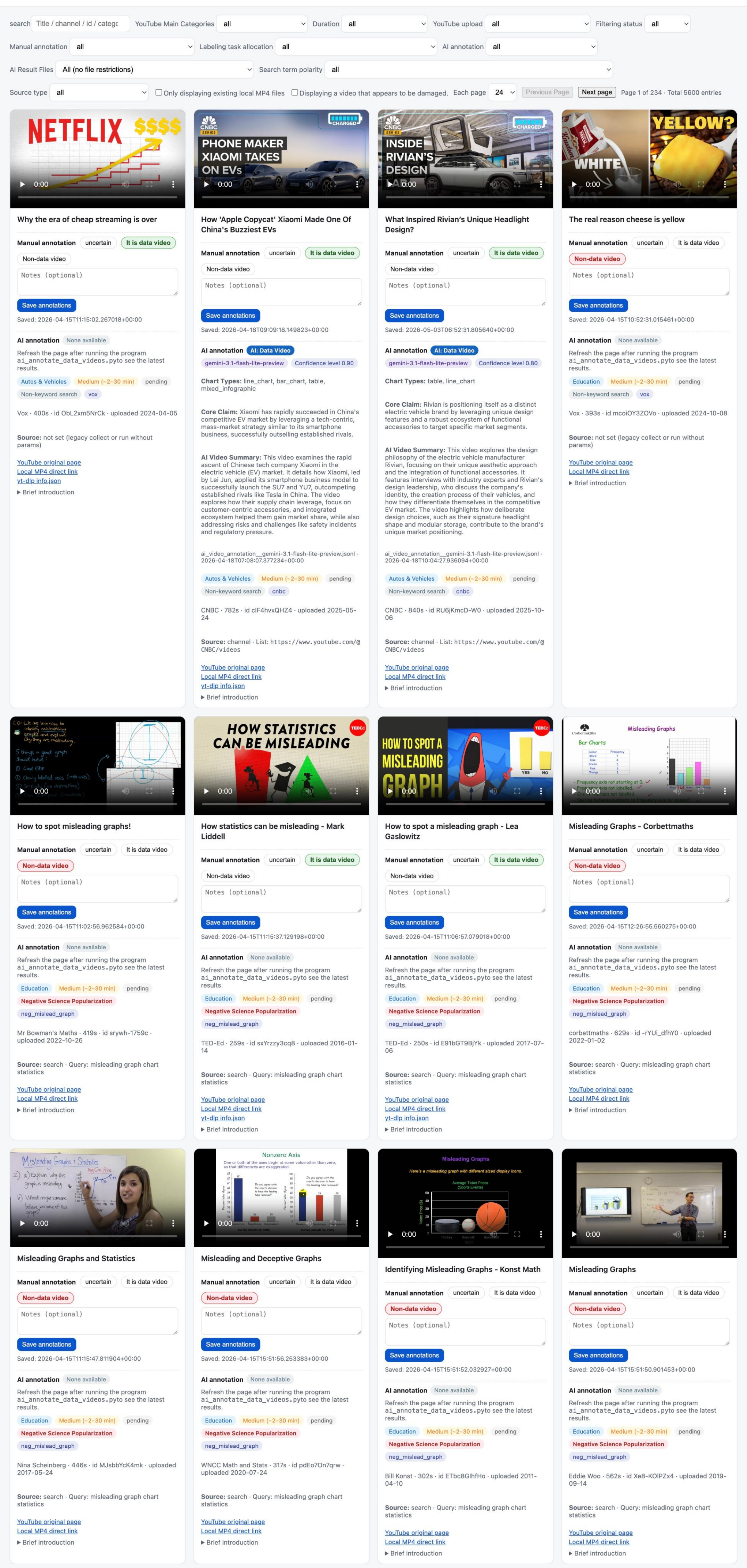}
    \caption{The human review interface for video screening. Annotators can watch the video, view AI-generated annotations, and confirm or correct the data video classification.}
    \label{fig:video_collect}
\end{figure}

Data video identification follows a two-stage pipeline of AI pre-screening and human verification. In the AI pre-screening stage, we feed video frames, subtitle text, and metadata (title, channel, duration, tags) into Gemini-3-Flash, which judges whether the video qualifies as a data video according to the three inclusion criteria in Section~\ref{sec:collection} and outputs structured annotations: if the judgment is negative, the model provides a specific rejection reason (e.g., no visualization, decorative only, live-action dominant, narration dominant); if positive, it further outputs identified visualization types, narrative stage annotations, language prior risk assessment, and key timestamps supporting the judgment. \rev{Annotators watch each video in the review interface (Figure~\ref{fig:video_collect}), verify the AI classification, and correct it where necessary.}

The complete screening prompt is shown below:

\begin{examplebox}{AI Screening Prompt for Data Video Identification}
You are an expert annotator for the DataVista data-video benchmark. Your task is to analyze a YouTube video and return a single JSON object with EXACTLY the schema below.

\textbf{Definition of ``data video'':}
A video is a DATA VIDEO if and only if ALL three conditions hold:
\begin{enumerate}
    \item It contains at least one visible data visualization (chart, graph, statistical map, data table, etc.) backed by real quantitative data.
    \item The visualization is CENTRAL and SUSTAINED---it must drive the narrative for a meaningful portion of the video, not appear as a brief location marker, evidence overlay, or decorative insert ($<$5\,s total is insufficient).
    \item The primary storytelling mode is DATA-DRIVEN: the charts/graphs are what the audience is meant to read and reason about.
\end{enumerate}

\textbf{CRITICAL rejection criteria} --- mark ``no'' if ANY of the following is true:
\begin{itemize}
    \item The video is mainly live-action footage, interviews, or on-the-ground reporting, even if it briefly shows maps or text overlays to locate events.
    \item The only ``visualizations'' are annotated photos, satellite imagery markups, crime-scene diagrams, or location/timeline graphics used as EVIDENCE in investigative journalism (not as quantitative data presentations).
    \item The main insight is delivered by narration/voiceover; charts merely illustrate what is already stated in words (narration-dominant).
    \item The video uses infographic-style text cards or flow diagrams to explain a PROCESS or POLICY (not to present statistical data trends or comparisons).
    \item The visualization count is 1--2 and they appear only briefly.
\end{itemize}

\textbf{Output schema} (abbreviated; full schema includes all fields below):
\begin{itemize}
    \item \texttt{is\_data\_video}: ``yes'' $|$ ``no''
    \item \texttt{rejection\_reason} (if ``no''): one of \{no\_visualization, decorative\_only, live\_action\_only, investigative\_journalism, narration\_dominant, process\_explainer, advertisement, pure\_text\_animation, other\}
    \item \texttt{visualization\_types}: list from 18 normalized types (bar\_chart, line\_chart, area\_chart, pie\_donut\_chart, scatter\_plot, bubble\_chart, map\_choropleth, map\_symbol, treemap, heatmap, table, flow\_diagram, network\_graph, timeline\_visual, pictograph\_isotype, annotated\_image, mixed\_infographic, other\_chart)
    \item \texttt{visualization\_count}: ``1-2'' $|$ ``3-5'' $|$ ``6-10'' $|$ ``10+''
    \item \texttt{visualization\_dynamic}: ``static'' $|$ ``animated'' $|$ ``mixed''
    \item \texttt{narrative\_setting}, \texttt{narrative\_rising\_climax}, \texttt{narrative\_resolution}: labels from a predefined narrative design space (e.g., statistic\_hook, showing\_contrast, recap)
    \item \texttt{data\_flow}: primary data-flow strategies (e.g., comparison, ranking, time\_series\_progression, cause\_and\_effect)
    \item \texttt{visual\_communication}: main visual-communication intents (e.g., emphasize, compare, reveal, summarize)
    \item \texttt{language\_prior\_risk}: ``low'' $|$ ``medium'' $|$ ``high'' --- estimates how much key data insight can be obtained from transcript alone
    \item \texttt{evidence\_timestamps}: 2--4 key timestamps with brief notes supporting the classification
    \item \texttt{confidence}: 0.0--1.0
\end{itemize}

The model receives the video (base64-encoded), the subtitle transcript when one is available, and metadata (title, channel, duration, tags) as input.
\end{examplebox}

\subsection{\rev{Structured Pre-extraction}}
\label{app:pre_extraction}

\rev{Videos that pass screening enter structured pre-extraction~\citep{zhang2026datamosaic,zhang2026datamosaicdemo}, which compiles a timestamped evidence index of each retained video to support question design. Gemini-3-Flash receives the full video and returns chart segments, narrative stages, argument chapters, and a video digest, with exact time fields required throughout so that annotators can jump to the corresponding segment in the review interface.}

\begin{examplebox}{\rev{Structured Pre-extraction Prompt}}
\rev{You are preparing an evidence map for the DataVista data-video benchmark. Watch the full video and return a single JSON object with EXACTLY the fields below. Enumerate every distinct data visualization, partition the video into narrative stages, and surface the moments worth examining. Timestamps must be exact.}

\medskip
\rev{\textbf{Output schema:}}
\begin{itemize}
    \item \rev{\texttt{video\_description}: a 2--4 sentence summary of the entire video.}
    \item \rev{\texttt{chart\_segments}: every distinct data visualization in the video, each with a stable \texttt{id} (\texttt{c1}, \texttt{c2}, \ldots), a \texttt{label} describing the chart form and its content, a \texttt{question\_reference} phrase safe to reuse in question stems, and a \texttt{time\_range} in \texttt{MM:SS-MM:SS}.}
    \item \rev{\texttt{narrative\_stages}: the time ranges of \texttt{setting}, \texttt{rising\_climax}, and \texttt{resolution}, together with a content-specific reference phrase for each stage.}
    \item \rev{\texttt{argument\_chapters}: 3--8 data-argument units subdividing the narrative stages, each with a \texttt{time\_range}, its parent \texttt{stage}, a \texttt{question\_reference} phrase, and the \texttt{key\_chart\_ids} anchoring that unit. Required when the video contains at least four distinct chart segments or exceeds three minutes.}
    \item \rev{\texttt{video\_digest}: a reviewer-facing briefing containing \texttt{distinct\_viz\_count} and a one-line inventory of the chart families present; \texttt{quiz\_hooks}, listing examinable moments with a \texttt{time\_hint}, the associated \texttt{chart\_ids}, and \texttt{suggested\_levels}; \texttt{cross\_modal\_hooks}, recording moments where a narration claim can be compared against the value the chart actually shows; and \texttt{stage\_notes}, summarizing the visualization activity within each stage.}
\end{itemize}

\medskip
\rev{\textbf{Consistency requirements:} \texttt{distinct\_viz\_count} must equal the length of \texttt{chart\_segments}; every \texttt{chart\_ids} entry must reference a declared segment id; \texttt{video\_digest} must remain consistent with the segments, stages, and chapters.}
\end{examplebox}

\rev{Annotators then verify this output against the video segment by segment, correcting inaccurate time ranges and chart descriptions and supplementing omitted visualizations, before the metadata is used for question design.}

\subsection{\rev{Question Screening and Quality Review}}
\label{app:question_screening}

\subsubsection{\rev{Text-Only Filtering}}
\label{app:text_only}

\paragraph{\rev{Screening rule.}}
\rev{We use Gemini-3-Flash for single-round text-only answering of candidate questions. The input contains question stems and options, without video frames, subtitles, audio, or reference answers. Predictions are matched against the reference letter for single-answer questions and the complete reference set for multiple-select questions. Matching candidates are filtered out; the rest proceed to human review.}

\begin{examplebox}{Text-Only Screening Prompt}
You are an answering module for benchmark screening. You are given a list of multiple-choice questions with NO video or image input. Answer each question using only the question text and options.

\medskip
\textbf{Hard constraints:}
\begin{itemize}[nosep]
    \item Output ONLY a JSON array of answer strings, one per question, in the same order as the input.
    \item For single-choice questions: one uppercase letter (A/B/C/D).
    \item For multi-select questions: \rev{all applicable uppercase letters in alphabetical order, comma-separated, no spaces (e.g., ``A,C''). The number of letters is not fixed.}
    \item No explanation, no extra fields, no markdown. Just the JSON array.
\end{itemize}

\medskip
Example output for 3 questions: {[}``B'', ``A,C'', ``D''{]}

\medskip
Questions:\\
Q1. \{question stem\}\\
A. \{option A\} ~~B. \{option B\} ~~C. \{option C\} ~~D. \{option D\}

\medskip
...

\medskip
Output (JSON array only):
\end{examplebox}

\paragraph{\rev{Screening statistics.}}
\rev{Of the 28{,}897 screened candidates, 21{,}804 (75.5\%) are filtered out and 7{,}093 proceed to human review. Table~\ref{tab:question_screening} reports the results by capability level.}

\begin{table}[t!]
\centering
\caption{\rev{Text-only screening results by capability level.}}
\label{tab:question_screening}
\ifshowrevisions\color{blue}\fi
\begin{tabular}{lrrrr}
\toprule
Level & Screened & Filtered & Sent to review & Filtered (\%) \\
\midrule
L1 & 9{,}306 & 6{,}369 & 2{,}937 & 68.4 \\
L2 & 9,876 & 7,845 & 2,031 & 79.4 \\
L3 & 9,715 & 7,590 & 2,125 & 78.1 \\
\midrule
Total & 28{,}897 & 21{,}804 & 7{,}093 & 75.5 \\
\bottomrule
\end{tabular}
\end{table}

\begin{samepage}
\paragraph{\rev{Cross-model filtering analysis.}}
\label{app:filter_cross_model}
\rev{To examine whether the filtering effect extends to other models, we sample equally sized sets of retained and filtered-out questions. We prioritize pairs from the same video with matching capability level, question type, and answer format, relaxing these criteria in a predefined order when necessary. Gemini-3.1-Pro, GPT-5, Claude-Sonnet-4.6, Qwen3-VL-32B-Instruct, and InternVL3.5-38B receive only question stems and options and are scored by exact match. We estimate 95\% confidence intervals for the filtered-out-minus-retained accuracy differences using 10,000 video-cluster bootstrap resamples.}
\par\end{samepage}

\begin{table}[t!]
\centering
\caption{\rev{Text-only accuracy (\%) on retained and filtered-out questions. Differences are filtered-out minus retained, in percentage points, with 95\% video-cluster bootstrap confidence intervals.}}
\label{tab:text_filter_cross_model}
\begingroup
\ifshowrevisions\color{blue}\fi
\small
\setlength{\tabcolsep}{4pt}
\renewcommand{\arraystretch}{1.15}
\begin{tabular}{@{}lrrc@{}}
\toprule
Model & Retained & Filtered out & Difference [95\% CI] \\
\midrule
Gemini-3.1-Pro & 55.6 & 85.6 & +30.0 [25.2, 35.1] \\
GPT-5 & 51.5 & 82.0 & +30.5 [23.6, 37.5] \\
Claude-Sonnet-4.6 & 43.9 & 76.0 & +32.2 [24.9, 39.5] \\
Qwen3-VL-32B-Instruct & 36.5 & 60.0 & +23.4 [15.0, 31.8] \\
InternVL3.5-38B & 42.5 & 65.4 & +22.9 [16.6, 29.3] \\
\bottomrule
\end{tabular}
\endgroup
\end{table}

\rev{As shown in Table~\ref{tab:text_filter_cross_model}, text-only accuracy is 22.9--32.2 percentage points higher on filtered-out questions across all five models. This consistent difference indicates that the filtered-out questions are more readily answerable from question stems and options across model families.}

\subsubsection{\rev{Human Review}}
\label{app:question_review}

\paragraph{\rev{Reviewers and procedure.}}
\rev{Four reviewers who did not participate in question design review the candidates retained after text-only filtering. Tasks are assigned by video, and each question is checked separately by two reviewers against the original video for clear wording, video-supported reference answers, and plausible distractors.}

\paragraph{\rev{Level-specific criteria.}}
\rev{L1 checks numerical values, units, and the corresponding data entities. L2 checks the completeness of evidence across charts and time points, and the validity of comparisons, calculations, and support relations. L3 checks the evidence underlying narrative and visual-design judgments, including whether alternative answers are equally reasonable.}

\paragraph{\rev{Disagreement resolution and outcomes.}}
\rev{Reviewers resolve disagreements through discussion grounded in the original video evidence. Questions requiring revision are edited and their answers and options rechecked; those with unresolved ambiguity are removed. Among candidates retained after text-only filtering, 80.1\% pass review, 15.4\% pass after manual revision, and 4.5\% are rejected.}

\subsection{Video Source Distribution}
\label{app:source_dist}

We summarize the most prolific source channels within each topic domain. In the Economy domain, CNBC is the largest contributor (\rev{153} videos), followed by Half as Interesting (18) and Financial Times (16). The Society domain draws primarily from Pew Research Center (53), Happy Statistics (15), and Our World in Data (\rev{6}). The Science domain is mainly from popular science channels such as Two Minute Papers and Kurzgesagt -- In a Nutshell. The Politics domain concentrates on Vox (\rev{73}) and Bloomberg Television (\rev{19}), while the Culture domain features independent creator channels focused on statistics and rankings, such as World Statistic and BarChart Race. Overall, the source channels span four broad categories: professional media outlets, research institutions, educational science-communication channels, and independent data creators, ensuring diversity in both content expertise and visual style.

\subsection{Topic and Subcategory Distribution}
\label{app:topic_dist}
Table~\ref{tab:topic-distribution} shows the two-level topic taxonomy of \benchmark, comprising five top-level domains further divided into 22 subcategories. The video counts for the five domains are: Economy (\rev{404}), Culture (173), Politics (\rev{152}), Society (145), and Science (87). \rev{Economy is the largest domain, accounting for 42.0\% of videos. This is consistent with economic reporting's emphasis on quantitative indicators, trends, and comparisons, which lends itself to video narratives organized around data visualizations~\citep{bu2026newsvis}.} At the subcategory level, \textit{Econ: Market Dashboard} (\rev{177}), \textit{Econ: Macro Comparison} (115), and \textit{Culture: Ranking Race} (105) are the three largest subcategories, whereas \textit{Politics: Event \& Governance} (4) and \textit{Science: Educational Explainer} (5) are among the smallest. \rev{Per-domain evaluation and equally weighted domain results are provided in Appendix~\ref{app:topic_breakdown}.}

\input{table/theme_distribution}

\subsection{Video Duration Distribution}
\label{app:duration_dist}

\begin{figure}[t]
    \centering \includegraphics[width=\textwidth]{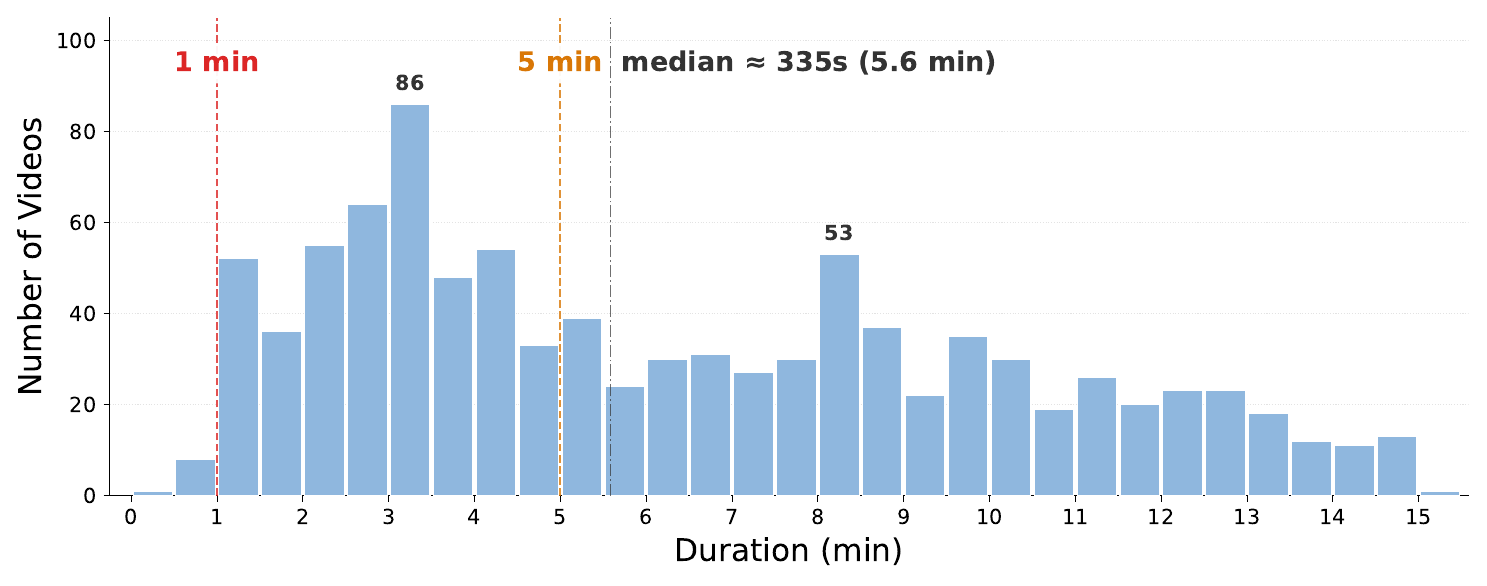}
    \caption{Distribution of video durations (30-second bins).}
    \label{fig:duration_dist}
\end{figure}
 
Figure~\ref{fig:duration_dist} shows the histogram of video durations (binned at 30-second intervals). The median duration is approximately 335 seconds ($\approx$5.6 minutes), with about \rev{45.1\%} of videos falling within the 60--300 second (1--5 minute) range. The shortest video is 28 seconds and the longest is 900 seconds (15 minutes). The distribution is not unimodal and symmetric: the primary peak is located at the 180--210 second (3--3.5 minute) bin, but secondary bumps appear near 240 seconds and 480 seconds. This reflects the coexistence of two narrative modes: short commentary clips and medium-length explanatory data videos. This distribution is consistent with observations by \citet{amini2015understanding} and \citet{DBLP:journals/tvcg/YangXLLGSQC22} regarding the typical narrative density of professional data videos.

\subsection{Video View Count Distribution}
\label{app:viewcount_dist}

\begin{figure}[t]
    \centering \includegraphics[width=\textwidth]{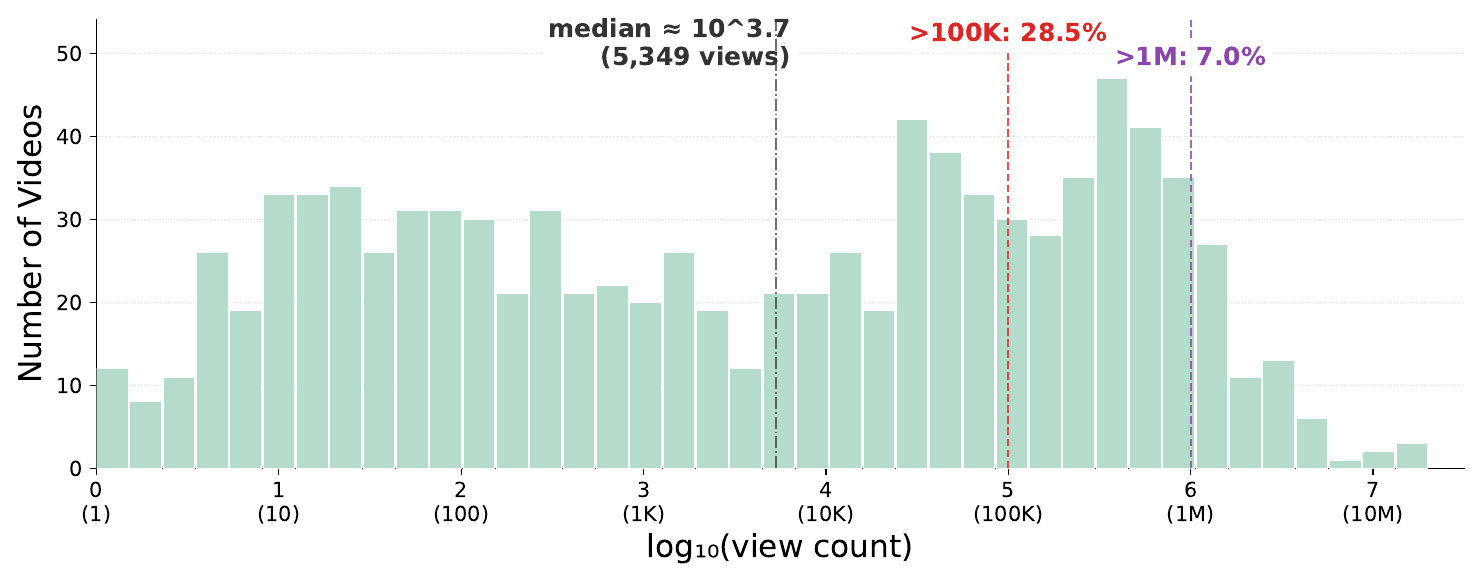}
    \caption{Distribution of video view counts ($\log_{10}$ scale).}
    \label{fig:viewcount_dist}
\end{figure}
 
Figure~\ref{fig:viewcount_dist} presents the $\log_{10}$-scaled histogram of video view counts (40 equal-width bins; \rev{945} videos have view count records). The median view count is approximately \rev{5{,}349}. About \rev{28.5\%} of videos have more than 100K views, and approximately \rev{7.0\%} exceed 1M views. The distribution exhibits a bimodal pattern: the first peak is located near $\log_{10} \approx 1.6$ ($\approx$40 views), corresponding to recently collected videos still in their cold-start phase; the second, broader bump spans $\log_{10} \approx 2$--$5$ (thousands to hundreds of thousands), corresponding to the typical reach of mainstream professional media content. This bimodal structure corroborates \benchmark's simultaneous attention to both \emph{timeliness} and \emph{influence} during data collection.

\subsection{Publication Date Distribution}
\label{app:pubdate_dist}

\begin{figure}[t]
    \centering \includegraphics[width=\textwidth]{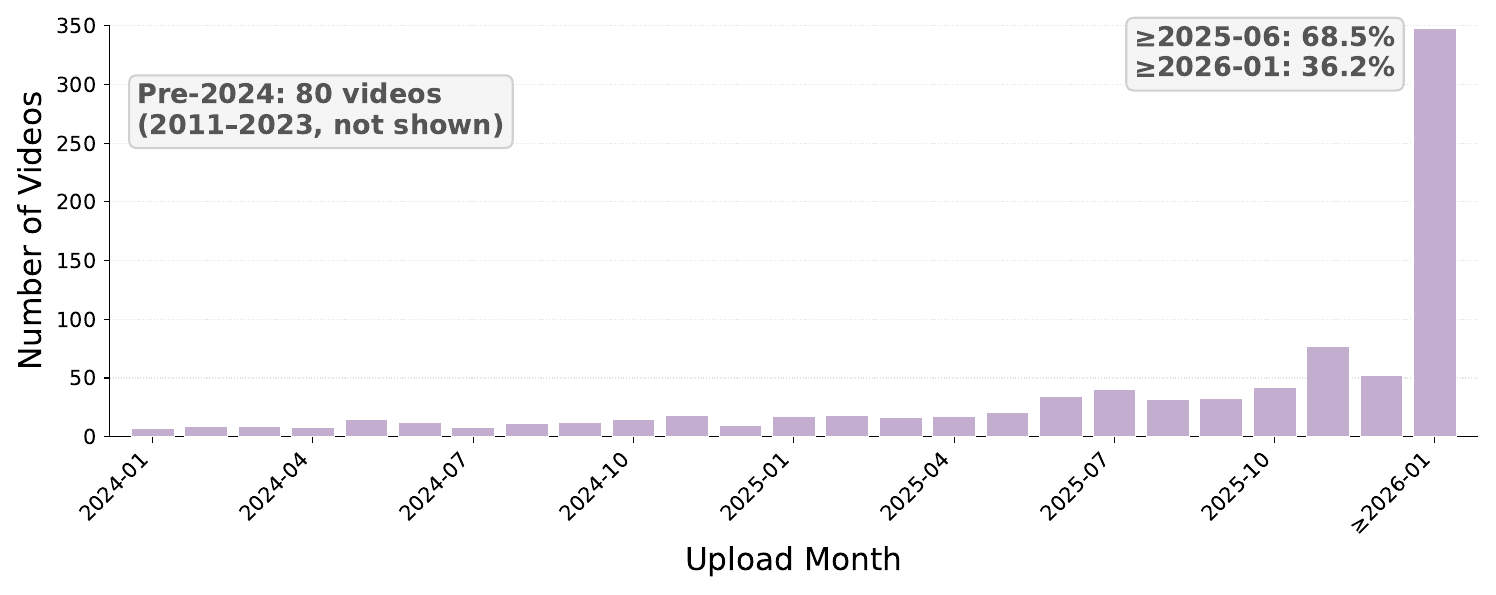}
    \caption{Monthly distribution of video upload dates.}
    \label{fig:pubdate_dist}
\end{figure}
 
Figure~\ref{fig:pubdate_dist} shows the monthly distribution of video publication dates. Approximately \rev{68.5\%} of videos were published in or after June 2025; by calendar year, \rev{399 videos (41.5\%)} were collected from 2025 and \rev{348 videos (36.2\%) were published in or after January 2026}. A ramp-up trend is visible from mid-2024 through the beginning of 2026, reflecting the dataset's deliberate strategy of skewing toward more recent content to reduce data contamination risk(see Appendix~\ref{app:contamination} for details).

\subsection{Chart Type Distribution}
\label{app:charttype_dist}

\input{table/chart_types}

Table~\ref{tab:charttype_dist} summarizes the distribution of chart types. We annotated each video with structured visualization type labels; videos containing multiple chart types receive separate entries for each type, so the total number of entries can exceed the number of videos. A total of 18 normalized chart types were identified, yielding \rev{2{,}210} entries across \rev{961} videos (an average of $\approx$2.30 types per video). The distribution is highly concentrated: bar charts appear in at least \rev{64.9\%} of videos (\rev{624} entries) and line charts in about \rev{42.6\%} (\rev{409} entries); bar, line, and area chart entries together account for approximately \rev{47.9\%} of all entries. Infographics ($\approx$12.0\% of entries), tables ($\approx$10.3\%), and two map variants (choropleth maps at $\approx$\rev{10.3\%} and symbol maps at $\approx$\rev{10.0\%}, both measured by video coverage rate) also appear frequently. Pie/donut charts, timelines, pictograms, and flow diagrams each fall in the moderate range. Scatter plots, bubble charts, heatmaps, treemaps, and network diagrams belong to the long tail, collectively covering fewer than about 10\% of videos when counted by whether at least one such type appears, yet they serve as important probes for model comprehension on less conventional chart types. This distribution aligns well with the content characteristics of professional data videos, which emphasize temporal and categorical comparisons.

\subsection{Question Count Distribution}
\label{app:question_dist}

\rev{After text-only filtering and human review, \benchmark retains 6{,}775 questions covering 961 videos. The median number of questions per video is 7, with a mean of 7.0, a range of 1--19, and first and third quartiles of 5 and 9. Videos with 5--10 questions account for 63.1\% of the dataset; those with 1--4, 11--15, and at least 16 questions account for 22.3\%, 13.4\%, and 1.2\%, respectively.}

\rev{L1, L2, and L3 contain 2{,}817, 1{,}908, and 2{,}050 questions, respectively, with per-video medians of 3, 2, and 2. The three levels cover 93.5\%, 83.4\%, and 88.4\% of videos, respectively, so most videos have questions at each level.}

%% file: table/theme_distribution.tex
\begin{table}[t!]
\centering
\caption{Distribution of 5 major themes and 22 subclass topics in \benchmark.}
\label{tab:topic-distribution}
\small
\begingroup
\ifshowrevisions\color{blue}\fi
\begin{tabular}{llrc}
\toprule
\textbf{Theme} & \textbf{Subclass Topic} & \textbf{Videos} & \textbf{Pct.(\%)} \\
\midrule
Economy (404) & Market Dashboard        & 177 & 18.4 \\
              & Macro Comparison        & 115 & 12.0 \\
              & Earnings \& Equities    &  91 &  9.5 \\
              & Ranking \& Country Race &  16 &  1.7 \\
              & Other                   &   5 &  0.5 \\
\midrule
Culture (173) & Ranking Race            & 105 & 10.9 \\
              & Visual Storytelling     &  34 &  3.5 \\
              & Sports Analytics        &  15 &  1.6 \\
              & Film \& Entertainment   &  13 &  1.4 \\
              & Other                   &   6 &  0.6 \\
\midrule
Politics (152)& Narrative Explainer     &  77 &  8.0 \\
              & Mainstream News         &  39 &  4.1 \\
              & Other                   &  32 &  3.3 \\
              & Event \& Governance     &   4 &  0.4 \\
\midrule
Society (145) & Demographic Dynamics    &  67 &  7.0 \\
              & Survey \& Public Opinion&  52 &  5.4 \\
              & Relational Patterns     &  11 &  1.1 \\
              & Geo-social Analysis     &   9 &  0.9 \\
              & Other                   &   6 &  0.6 \\
\midrule
Science (87)  & Science Communication   &  75 &  7.8 \\
              & Educational Explainer   &   5 &  0.5 \\
              & Geospatial Analysis     &   7 &  0.7 \\
\midrule
\textbf{Total}& & \textbf{961} & \textbf{100.0} \\
\bottomrule
\end{tabular}
\endgroup
\end{table}

%% file: table/chart_types.tex
\begin{table*}[t!]
    \centering
    \small
    \setlength{\tabcolsep}{5.8pt}
    \caption{Distribution of chart types across \rev{961} videos (\rev{2{,}210} total entries; $\approx$2.30 per video). \textbf{Count} is the occurrence count for each type; \textbf{\%} is its share of all entries. ``Other'' subsumes miscellaneous and unclassified types.}
    \label{tab:charttype_dist}
    \begin{tabular}{lrr@{\hskip 0.8em}lrr@{\hskip 0.8em}lrr}
        \toprule
        \textbf{Chart Type} & \textbf{Count} & \textbf{\%}
        & \textbf{Chart Type} & \textbf{Count} & \textbf{\%}
        & \textbf{Chart Type} & \textbf{Count} & \textbf{\%} \\
        \midrule
        Bar Chart            & \rev{624} & 28.2 & Pictograph / Isotype &  59 & 2.7 & Bubble Chart    &  17 & 0.8 \\
        Line Chart           & \rev{409} & \rev{18.5} & Flow Diagram         &  59 & 2.7 & Other           &  16 & 0.7 \\
        Infographic          & 266 & 12.0 & Timeline             &  56 & 2.5 & Heatmap         &  15 & 0.7 \\
        Table                & \rev{228} & 10.3 & Annotated Image      &  \rev{42} & 1.9 & Treemap         &  11 & 0.5 \\
        Pie / Donut          & 112 &  \rev{5.1} & Scatter Plot         &  33 & 1.5 & Network Graph   &   6 & 0.3 \\
        Choropleth Map       &  99 &  4.5 & Area Chart           &  \rev{26} & 1.2 &                 &     &     \\
        Symbol Map           &  96 &  4.3 & Unknown              &  \rev{36} & \rev{1.6} &                 &     &     \\
        \midrule
        & & & & & & \textbf{Total} & \rev{\textbf{2{,}210}} & \textbf{100} \\
        \bottomrule
    \end{tabular}
\end{table*}

%% file: src/appendix-c-case-studies.tex
\section{Case Studies}
\label{appendix:case_studies}

This appendix presents representative questions from \benchmark as a supplement to Section~\ref{sec:experiment}. \rev{The cases are organized by capability level (L1, L2, and L3), using video evidence to explain what each question requires and to analyze differences in the answers of representative models.}

\rev{Each case presents the question and options, timestamped key frames, the basis for the answer, and model responses side by side, with correct answers highlighted in green. The analysis connects evidence to responses: we explain how the key frames support the answer, compare correct selections, omissions, and incorrect choices, and discuss how these differences affect numerical judgments, argument completeness, and design understanding.}

\noindent\begin{minipage}{\linewidth}
\subsection{Level 1: Data Perception}
\label{app:case_l1}

\rev{This subsection examines model responses on numerical comparison and visual encoding recognition through quarterly peak identification and legend mapping. The cases require comparing quarterly sales within a specified year and identifying the meaning of a line from its color and style, respectively. We analyze response differences using the displayed values, visual marks, and model selections.}
\end{minipage}

\paragraph{Case 1: Quarterly Peak Identification (L1, Data Fact Reading). }

\begin{figure*}[t!]
  \centering  \includegraphics[width=\textwidth]{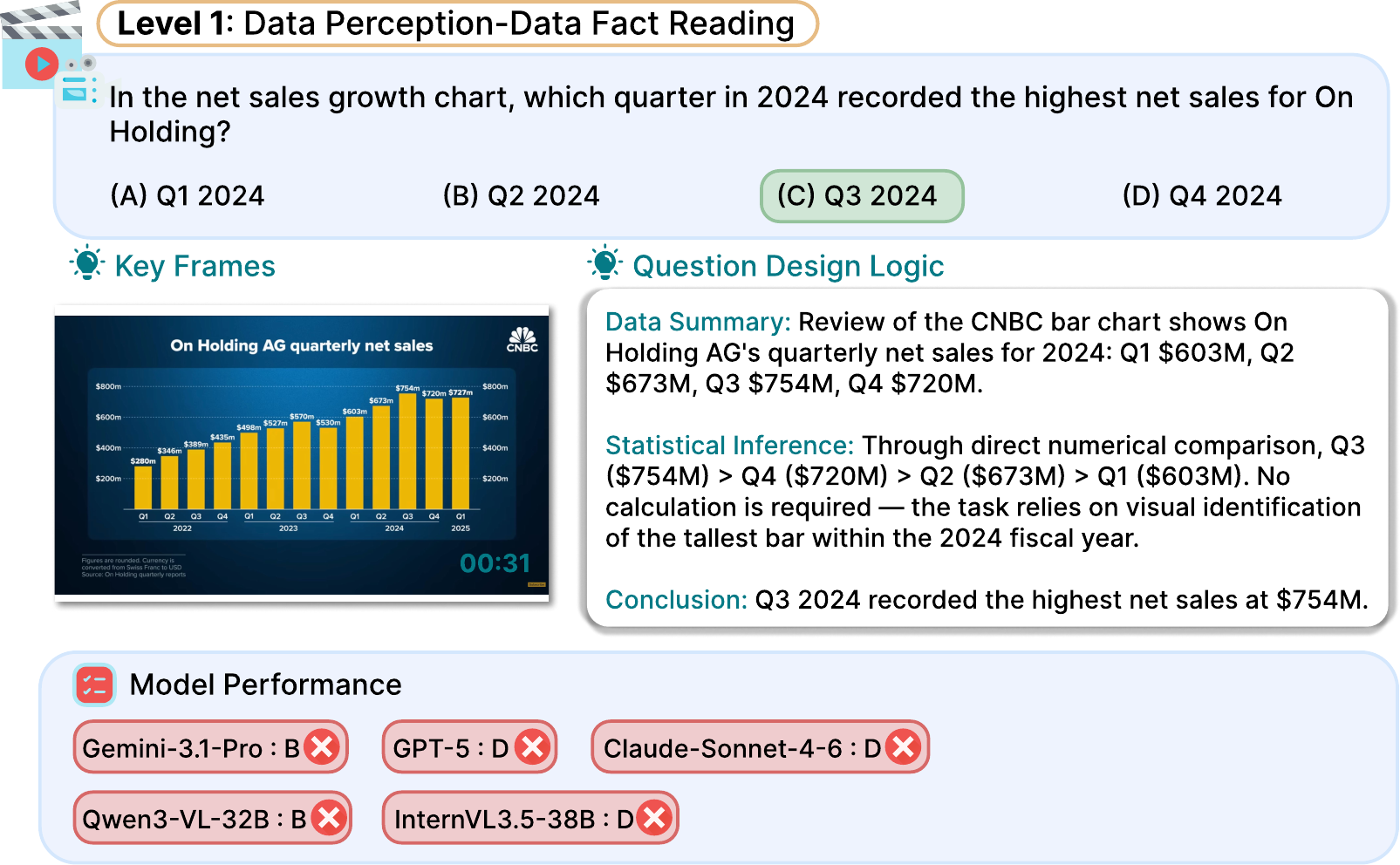}
  \caption{An example L1 question in \benchmark (Data Fact Reading)}
  \label{fig:app_case_l1_data_fact}
\end{figure*}

As shown in Figure~\ref{fig:app_case_l1_data_fact}, the question asks which quarter in 2024 recorded the highest net sales for On Holding. The correct answer (C, Q3 2024) can be determined by visual comparison of bar heights within the 2024 fiscal year, where Q3 ($754M$) is the tallest bar. However, all five representative models answer incorrectly: Gemini-3.1-Pro and Qwen3-VL-32B select Q2, while GPT-5, Claude-Sonnet-4.6, and InternVL3.5-38B select Q4. \rev{The fourth quarter, selected by three models, records sales of $720M$, approximately 4.5\% below the third quarter. This case illustrates that even a basic task such as identifying an annual peak can involve difficulty distinguishing closely valued alternatives, with models selecting the second-highest quarter as the highest.}

\paragraph{Case 2: Legend-Element Mapping (L1, Chart Element Recognition).}

\begin{figure*}[t!]
  \centering  \includegraphics[width=\textwidth]{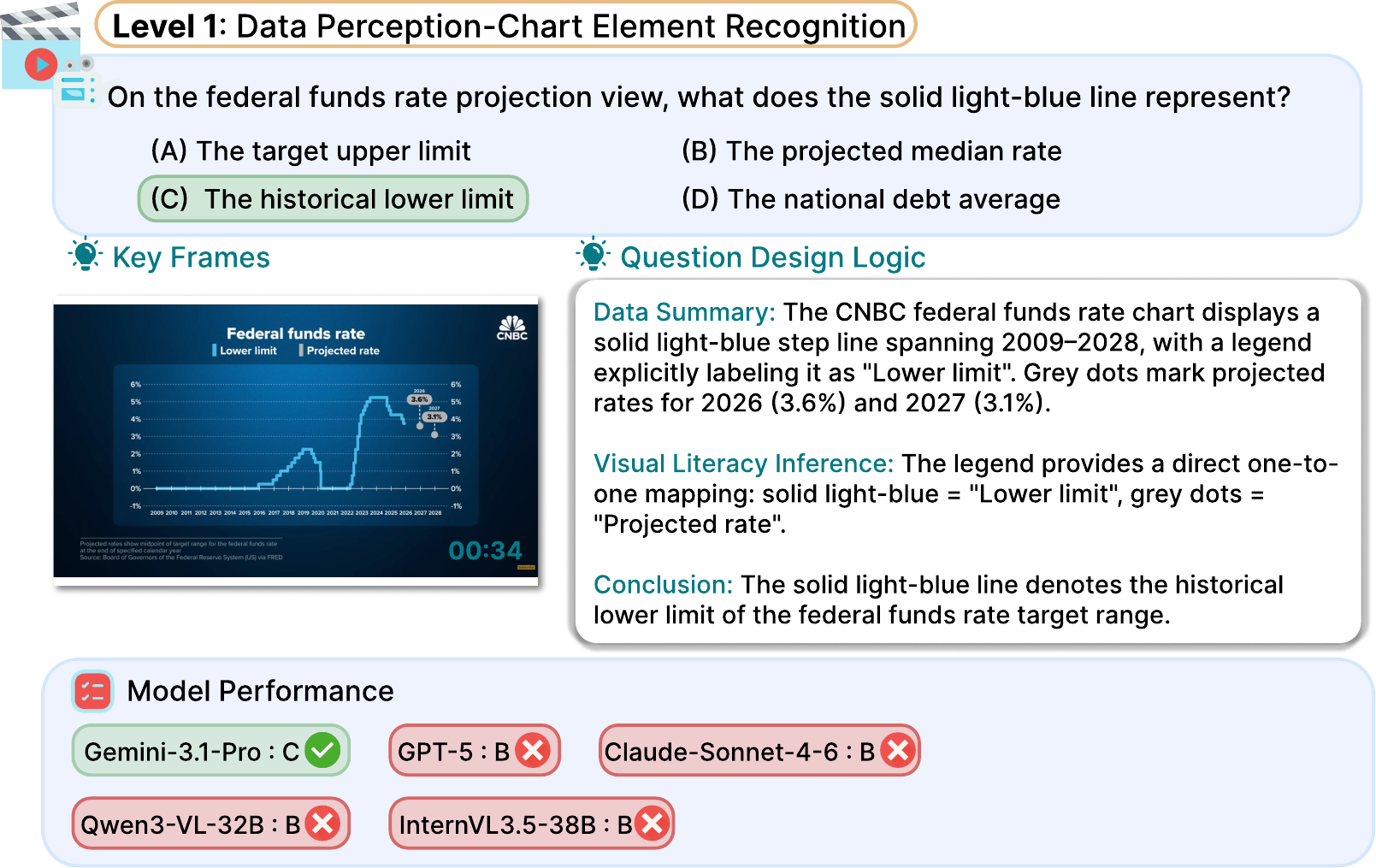}
  \caption{An example L1 question in \benchmark (Chart Element Recognition).}
\label{fig:app_case_l1_chart_element}
\end{figure*}

As shown in Figure~\ref{fig:app_case_l1_chart_element}, the question asks what the solid light-blue line represents on a federal funds rate projection chart. The correct answer (C, the historical lower limit) is directly readable from the chart legend, which explicitly maps the solid light-blue line to "Lower limit." Only Gemini-3.1-Pro answers correctly; all other models select B (the projected median rate), \rev{assigning the meaning of the grey projection dots to the solid light-blue line. Although the legend explicitly distinguishes the two types of marks, the four models fail to match the specified line to its meaning, revealing difficulty in recognizing the mapping between color, mark shape, and data meaning.}

\subsection{Level 2: Temporal Reasoning}
\label{app:case_l2}

\rev{This subsection uses argument synthesis and cross-temporal arithmetic cases to illustrate the requirements of cross-segment evidence integration in L2 temporal reasoning and the corresponding model performance. Argument synthesis requires selecting all key facts from different charts that support the central claim, while cross-temporal arithmetic requires reading data at specified times and calculating the change. The case analyses examine whether models select the complete set of supporting evidence and correctly perform numerical comparisons across time, respectively.}

\paragraph{Case 1: Cross-Chart Evidence Integration (L2, Argument Synthesis). }

\begin{figure*}[t!]
  \centering
  \includegraphics[width=\textwidth]{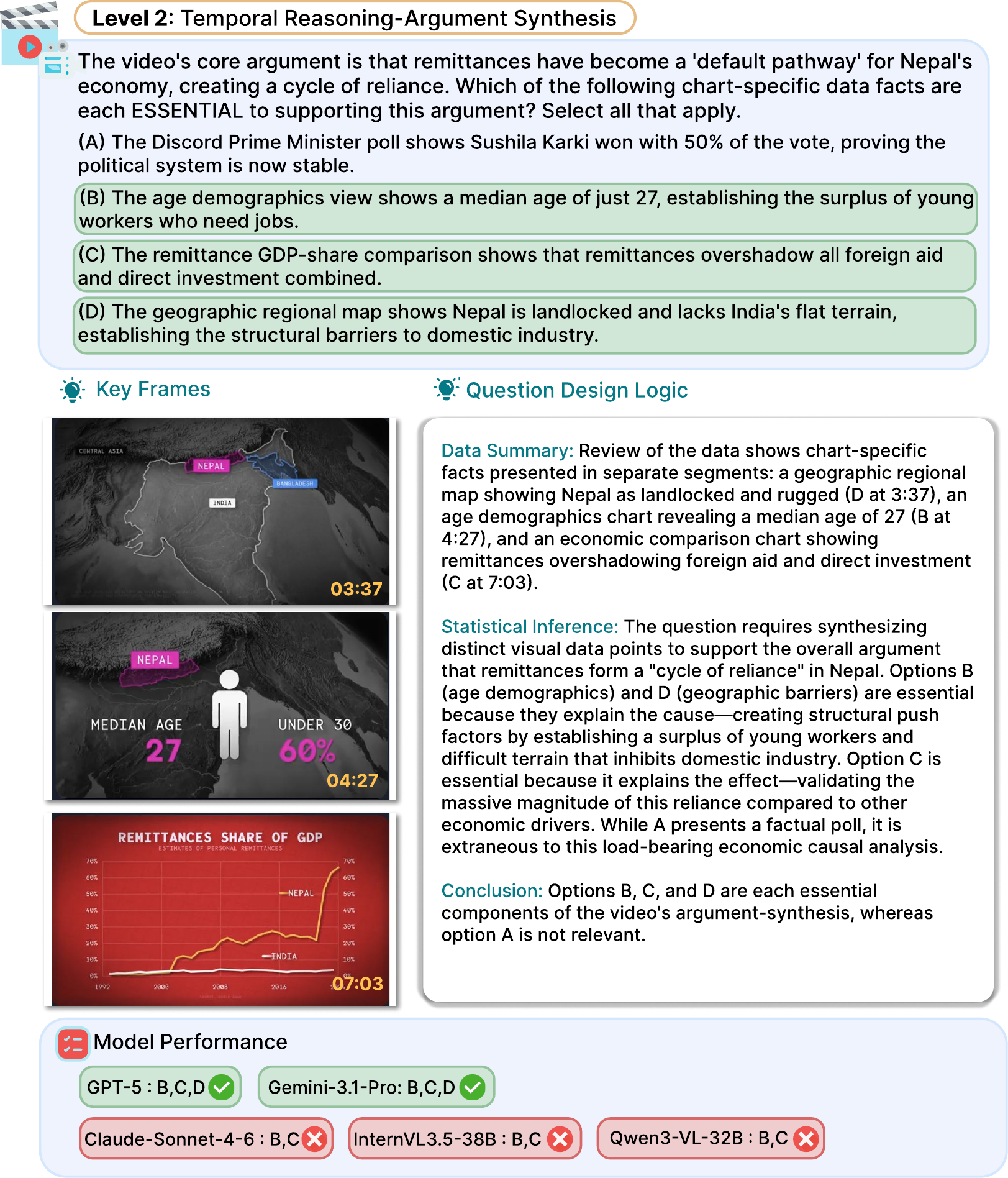}
  \caption{An example L2 question in \benchmark (Argument Synthesis).}
  \label{fig:case_l2_argument_synthesis}
\end{figure*}

Figure~\ref{fig:case_l2_argument_synthesis} shows the video argues that remittances have become a "default pathway" for Nepal's economy. The model must select all chart-specific data facts that are essential to supporting this argument from across three non-adjacent segments: a geographic map (03:37), an age demographics chart (04:27), and a remittance GDP-share comparison (07:03). The correct answer (B, C, D) requires synthesizing structural push factors (young workforce, landlocked terrain) with the economic outcome (remittance dominance), while option A (a Discord poll result) is factually present but irrelevant to the causal chain. GPT-5 and Gemini-3.1-Pro correctly identify all three essential pieces; \rev{the other three models select B and C but omit D. Options B and C concern the population structure and the economic importance of remittances, while D adds the geographic constraints on domestic industrial development. All three answers omit the explanatory link concerning limited domestic development opportunities, leaving the selected evidence incomplete in explaining why remittances have become a "default pathway" for Nepal's economy.}

\paragraph{Case 2: Cross-Temporal Arithmetic (L2, Cross-Temporal Change). }

\begin{figure*}[t!]
  \centering
  \includegraphics[width=\textwidth]{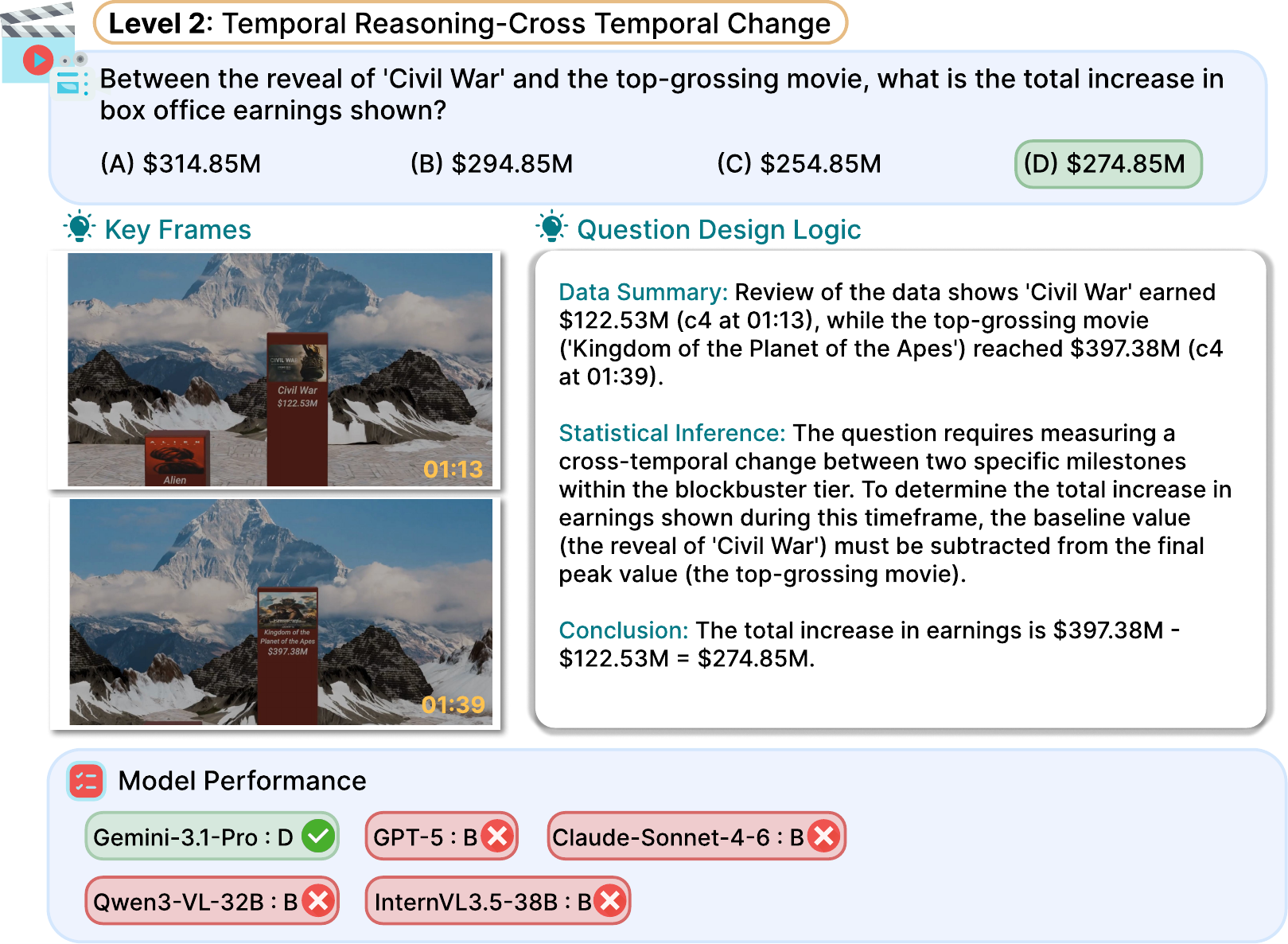}
  \caption{An example L2 question in \benchmark (Cross-Temporal Change). }
  \label{fig:case_l2_cross_temporal}
\end{figure*}

As shown in Figure~\ref{fig:case_l2_cross_temporal}, the question asks for the total increase in box office earnings between two specific milestones in an animated ranking. The correct answer (D, $274.85M$) requires subtracting the value at the first milestone ($122.53M$ at 01:13) from the peak value ($397.38M$ at 01:39). Only Gemini-3.1-Pro answers correctly; the remaining four models all select B ($294.85M$). \rev{This shared incorrect answer exceeds the correct increase by $20M$. The question combines basic subtraction with matching data across time: models must locate two specified moments in a continuously updated ranking, read the respective values, and calculate the change. All four models fail to obtain the correct result on this combined task, illustrating that basic numerical operations in dynamic videos also depend on accurate temporal localization and data reading.}

\subsection{Level 3: Narrative Understanding}
\label{app:case_l3}

\rev{This subsection uses cases of visual communication intent, counterfactual design, and {narrative structure} to illustrate the task requirements and model performance in L3 narrative understanding. The three cases examine whether models can identify the visual techniques used in a video, assess the benefits and costs of changing where information is presented, and understand how different charts are organized within a specified segment, respectively. The analyses focus on which design elements models omit or misidentify and how these answer discrepancies affect their interpretation of information emphasis, narrative pacing, and argument structure.}

\paragraph{Case 1: Visual Communication Technique Identification (L3, Visual Communication Intent).}

\begin{figure*}[t!]
  \centering
  \includegraphics[width=\textwidth]{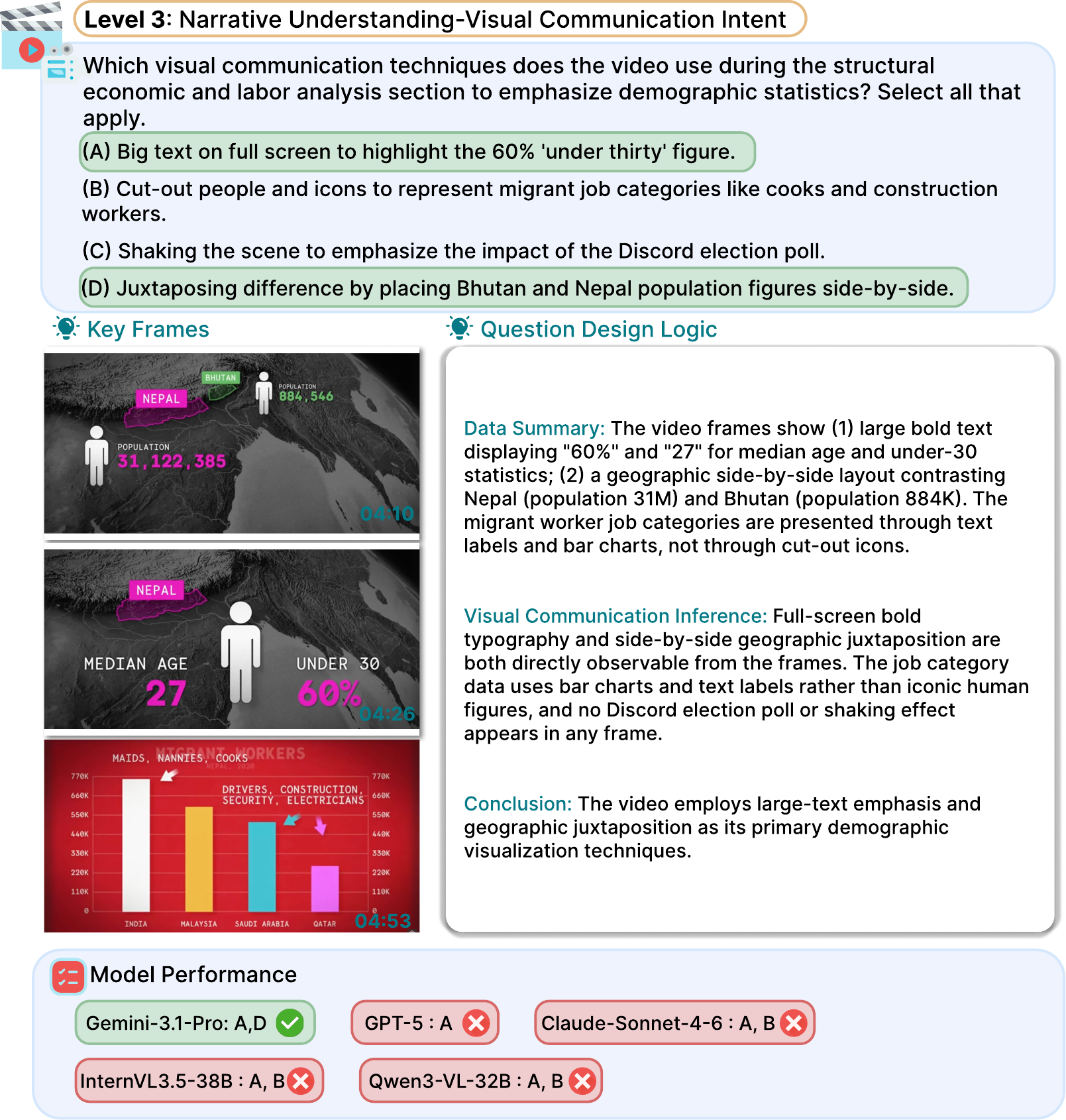}
  \caption{An example L3 question in \benchmark (Visual Communication Intent).}
  \label{fig:case_l3_visual_com}
\end{figure*}

As shown in Figure~\ref{fig:case_l3_visual_com}, the question asks which visual communication techniques the video uses to emphasize demographic statistics. The correct answer (A, D) involves recognizing full-screen bold typography for the "60\%" figure and a side-by-side geographic juxtaposition of Nepal and Bhutan populations. \rev{These techniques communicate information by emphasizing an individual value and establishing a comparison between countries, respectively.} Only Gemini-3.1-Pro correctly identifies both techniques. \rev{GPT-5 selects only A, omitting the juxtaposition in D; Claude-Sonnet-4.6, Qwen3-VL-32B, and InternVL3.5-38B select A and B, both omitting D and incorrectly selecting the cut-out people and icons in B. However, the job categories are presented through bar charts and text labels, without these human figures. All models identify the emphasis on an individual value, but all four incorrect answers omit the comparison established through spatial juxtaposition, and three also attribute absent graphical elements to the video design. These responses reflect incomplete understanding of visual communication: they miss the comparative relationship conveyed by the layout and include design judgments unsupported by the displayed content.}

\paragraph{Case 2: Counterfactual Design Reasoning (L3, Counterfactual Analysis).}

\begin{figure*}[t!]
  \centering  \includegraphics[width=\textwidth]{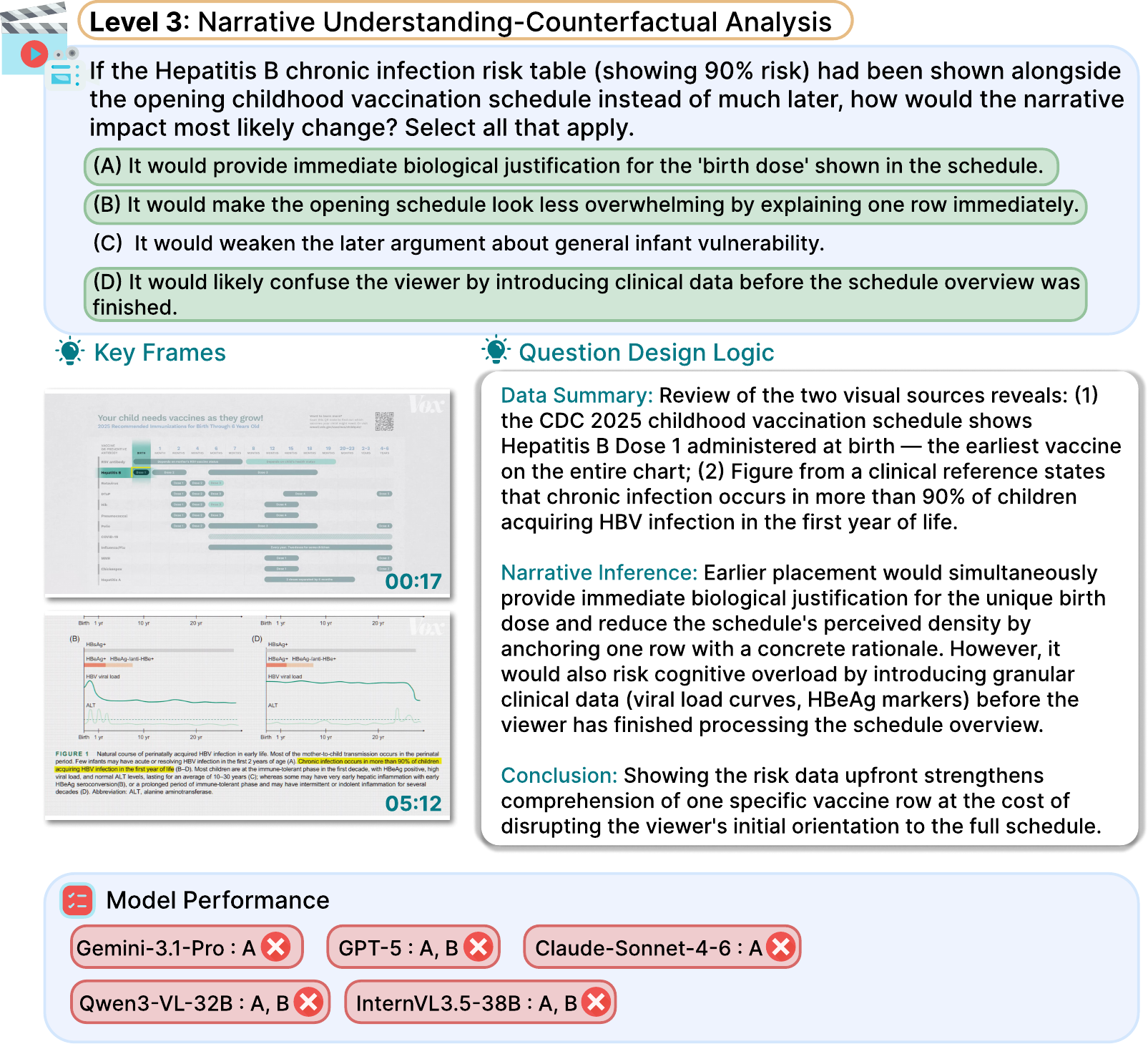}
  \caption{An example L3 question in \benchmark (Counterfactual Analysis). }
\label{fig:app_case_l3_counterfactual}
\end{figure*}

As shown in Figure~\ref{fig:app_case_l3_counterfactual}, \rev{this multiple-select question} asks how the narrative impact would change if a Hepatitis B risk table were shown alongside the opening vaccination schedule instead of later. \rev{The correct answers are A, B, and D: presenting risk data earlier helps explain vaccination at birth and makes the schedule easier to understand, but may also interrupt the opening overview and increase cognitive load.} No model gets this fully correct: Gemini-3.1-Pro and Claude-Sonnet-4.6 select only A, while the others select A and B but miss D. \rev{All five models recognize the explanatory role of the risk data but omit the potential information overload from introducing clinical details earlier. This shared omission indicates that their assessments of the design change in this case emphasize local explanatory benefits without fully accounting for its effect on the overall narrative pacing.}

\paragraph{Case 3: Organizing Evidence Across Charts (L3, {Narrative Structure}).}

\begin{figure*}[t!]
  \centering
  \includegraphics[width=\textwidth]{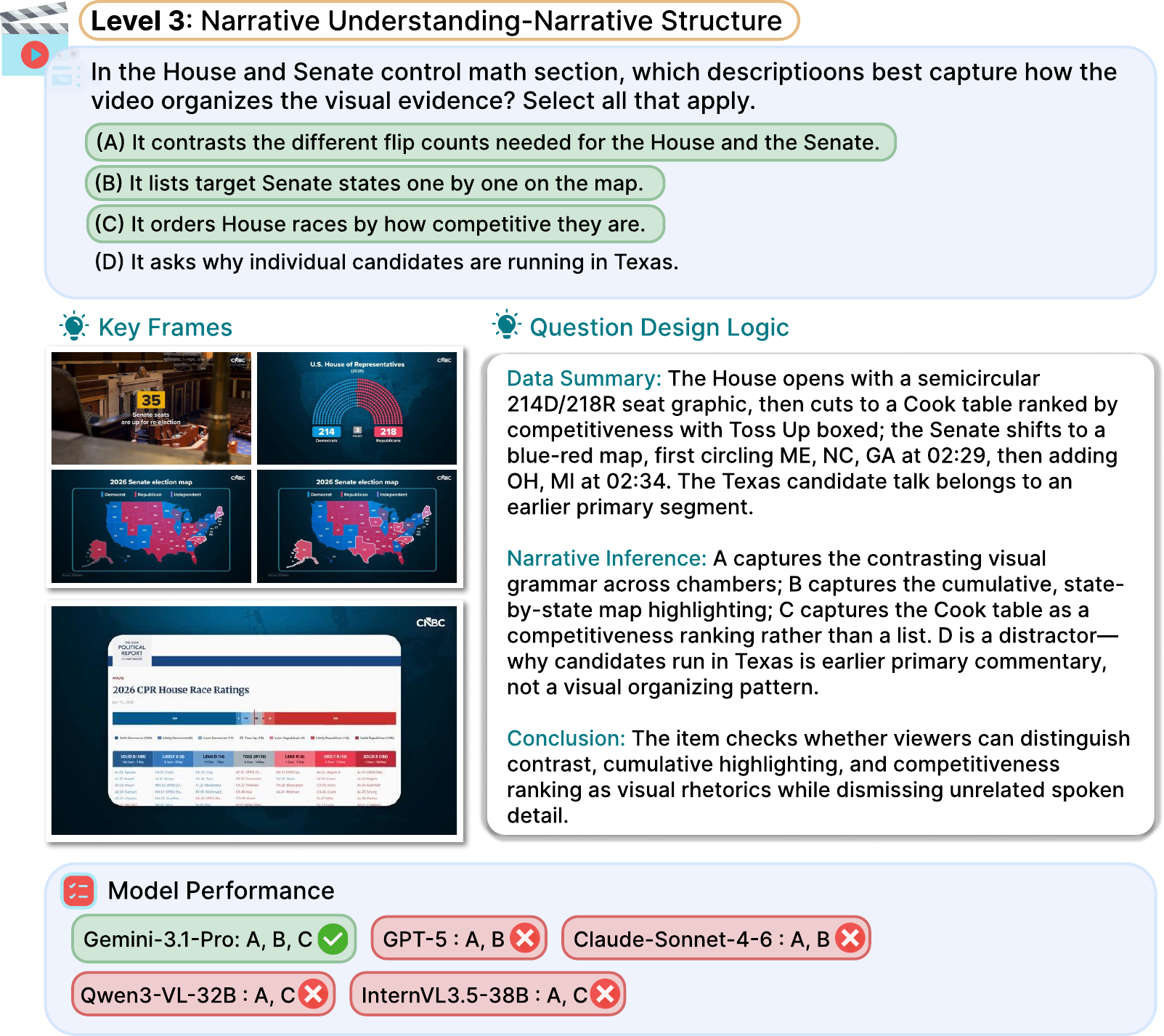}
  \caption{An example L3 question in \benchmark ({Narrative Structure}).}
  \label{fig:case_l3_narrative}
\end{figure*}

As shown in Figure~\ref{fig:case_l3_narrative}, the question asks which descriptions capture how the video organizes the visual evidence in the House and Senate control-math section. The correct answer is A, B, C. \rev{A corresponds to the two seat graphics: 35 Senate seats are up for re-election, and the House stands at 214 Democrats to 218 Republicans. B corresponds to target states marked successively on the map: Maine, North Carolina, and Georgia at 02:29, with Ohio and Michigan added at 02:34. C corresponds to the Cook Report table, which groups House races by competitiveness. D concerns the Texas candidates introduced at the opening and does not belong to this section. Only Gemini-3.1-Pro selects the full set A, B, C. GPT-5 and Claude-Sonnet-4.6 select A and B, omitting the Cook grouping; Qwen3-VL-32B and InternVL3.5-38B select A and C, omitting the states marked successively on the map. The first pair misses how the races are organized by competitiveness, while the second misses the progressive introduction of key states. All four models identify the seat comparison and exclude material from outside the section, but fail to fully recognize how the video combines comparison, progressive highlighting, and categorization to organize its account of the electoral landscape.}

%% file: src/appendix-d-evaluation-details.tex
\clearpage
\section{Evaluation Details}
\label{appendix:eval_details}

This appendix provides the detailed evaluation settings as a supplement to Section~\ref{sec:experiment}.

\subsection{\rev{Input Organization and Prompt Templates}}
\label{app:prompts}

\rev{The main evaluation uses shared answer requirements, with questions from the same video submitted in one request. Closed-source models are queried through official APIs; open-source models use official instruct/chat checkpoints. Decoding is deterministic with temperature 0. Below, we summarize these requirements and the input organization for the conditions without and with subtitles.}

\begin{examplebox}{\rev{Answer Requirements}}
\ifshowrevisions\color{blue}\fi
Answer every question using the supplied evidence. Return only a JSON array in question order, with one object per question containing \texttt{qid}, \texttt{answer}, \texttt{reason}, and \texttt{evidence}. Copy each question ID exactly; do not skip, merge, or reorder questions.
\begin{itemize}[nosep]
    \item \textbf{Answer:} For single-choice questions, return one letter: A, B, C, or D. For multiple-select questions, return comma-separated option letters, such as A,C. Every question requires an option answer.
    \item \textbf{Reason:} Provide the reason supporting the answer. Connect the cited observations to the answer and include key calculations when relevant. For multiple-select questions, explain which options satisfy the question and any important exclusions. Do not repeat the question.
    \item \textbf{Evidence:} Provide an array of source references, each containing \texttt{source\_id}, \texttt{timestamp}, and \texttt{observation}. Use only supplied source IDs and keep observations concise and specific to the cited source.
\end{itemize}
Do not use Markdown fences or add text outside the JSON array.
\end{examplebox}

\rev{The condition without subtitles provides 50 uniformly sampled video frames, each labeled with a source ID and timestamp. The subtitle condition adds subtitle text with source IDs and time information to the same frames. Frames and subtitles form the evidence list, followed by the question stems, question types, and options. Both conditions use the same answer requirements. Each question is labeled in the prompt as single-choice or multiple-select.}

\rev{Subtitles are obtained from YouTube. English subtitles are used for 921 videos, of which 386 have a creator-provided English track and 535 have only an automatically generated one. The remaining 40 videos have no English track and fall back to another language, most often Hindi.}

\begin{examplebox}{\rev{Input Organization}}
\ifshowrevisions\color{blue}\fi
\textbf{Available evidence:}\\
\{Frame source IDs and timestamps; corresponding images are attached.\}\\
\{Subtitle source IDs, time information, and text; included only with subtitles.\}
\medskip
\par\textbf{Question 1 | qid=\{question ID\} | SINGLE-CHOICE / MULTIPLE-SELECT}\\
\{question stem\}\\
A) \{option A\}\\
B) \{option B\}\\
C) \{option C\}\\
D) \{option D\}
\medskip
\par\ldots
\medskip
\par Now output the JSON array of exactly $N$ answer objects.
\end{examplebox}

\subsection{\rev{Random Selection and Most-Frequent Answers}}
\label{app:statistical_baselines}

\rev{We use two baselines without video content to provide random-guessing and answer-frequency references for model performance. Table~\ref{tab:statistical_baselines} reports the results.}

\paragraph{\rev{Random selection.}}
\rev{For single-choice questions, we select uniformly from the four options. For multiple-select questions, we select uniformly from the 15 nonempty option subsets. The exact-match probabilities are therefore $1/4$ and $1/15$, respectively. Averaging these per-question probabilities gives the expected accuracy of random selection.}

\paragraph{\rev{Most-frequent answers.}}
\rev{We partition videos into five folds and predict each fold using the most frequent answer in the other four folds. Frequencies are computed separately for single-choice and multiple-select questions, treating each complete option set as an answer for the latter. All questions from the same video remain in one fold. We pool predictions across the five folds to compute exact-match accuracy.}

\begin{table}[t!]
\centering
\caption{\rev{Exact-match accuracy (\%) of the statistical baselines. Random selection reports the theoretical expectation; most-frequent answers use video-level five-fold cross-validation.}}
\label{tab:statistical_baselines}
\begingroup
\ifshowrevisions\color{blue}\fi
\small
\begin{tabular*}{\linewidth}{@{\extracolsep{\fill}}lrrrr@{}}
\toprule
Baseline & L1 & L2 & L3 & Overall \\
\midrule
Random selection & 24.8 & 19.5 & 16.4 & 20.8 \\
Most-frequent answers & 30.8 & 18.3 & 31.5 & 27.5 \\
\bottomrule
\end{tabular*}
\endgroup
\end{table}

\subsection{\rev{Statistical Reliability}}
\label{app:statistical_reliability}

\rev{To account for dependence among questions from the same video, we perform 10,000 video-level bootstrap resamples with replacement (seed 42). Each resample retains all available responses from the selected videos and recomputes accuracy using the main-table scoring rule. The 2.5th and 97.5th percentiles form the 95\% confidence interval.}

\input{table/tab_accuracy_ci_projected}

\rev{As shown in Table~\ref{tab:accuracy_ci}, the interval half-widths for overall accuracy are approximately 1.2--1.4 percentage points. Gemini-3.1-Pro outperforms the best open-source model, Qwen3-VL-32B, by 12.7 percentage points, well beyond this margin, and the two 95\% intervals do not overlap, indicating that this gap reflects a genuine performance difference rather than sampling noise. The same resolution holds across the full capability range: overall accuracy among the 19 models spans 31.1\% for InternVL3-2B to 70.0\% for Gemini-3.1-Pro, a 38.9-percentage-point spread that is about 30 times the interval half-width, indicating that the benchmark yields discriminative estimates from the smallest open-source models to the strongest closed-source models.}

%% file: table/tab_accuracy_ci_projected.tex
\begin{table}[t!]
\centering
\caption{\rev{Exact-match accuracy (\%) and 95\% intervals.}}
\label{tab:accuracy_ci}
\begingroup
\ifshowrevisions\color{blue}\fi
\small
\setlength{\tabcolsep}{4pt}
\renewcommand{\arraystretch}{1.15}
\begin{tabular}{@{}lccc@{}}
\toprule
Model & w/o Sub. & w/ Sub. & Overall \\
\midrule
\multicolumn{4}{c}{\textit{Closed-Source Models}} \\
\midrule
Gemini-3.1-Pro & 68.2 [67.0, 69.7] & 71.7 [70.5, 73.1] & 70.0 [68.8, 71.3] \\
GPT-5 & 66.0 [64.5, 67.5] & 68.8 [67.5, 70.2] & 67.4 [66.1, 68.8] \\
Gemini-3-Flash & 64.1 [62.6, 65.7] & 68.5 [67.2, 70.0] & 66.3 [65.0, 67.7] \\
GPT-4o & 56.1 [54.9, 57.7] & 58.0 [56.8, 59.4] & 57.1 [56.0, 58.4] \\
Claude-Sonnet-4.6 & 61.3 [60.0, 62.8] & 65.8 [64.5, 67.2] & 63.6 [62.4, 64.9] \\
\midrule
\multicolumn{4}{c}{\textit{Open-Source Models}} \\
\midrule
Qwen3-VL-32B-Instruct & 54.7 [53.4, 56.4] & 59.8 [58.6, 61.4] & 57.3 [56.1, 58.8] \\
InternVL3.5-38B & 52.6 [51.3, 54.2] & 55.6 [54.3, 57.2] & 54.1 [52.9, 55.6] \\
InternVL3.5-14B & 45.2 [44.0, 46.7] & 49.6 [48.4, 51.2] & 47.4 [46.3, 48.9] \\
InternVL3.5-8B & 44.7 [43.5, 46.1] & 47.4 [46.2, 48.8] & 46.1 [44.9, 47.4] \\
Qwen3-VL-8B-Instruct & 44.1 [42.9, 45.7] & 48.2 [47.2, 49.8] & 46.2 [45.1, 47.6] \\
InternVL3-8B & 43.4 [42.2, 45.0] & 46.7 [45.5, 48.4] & 45.1 [43.9, 46.6] \\
Qwen2.5-VL-32B-Instruct & 54.0 [52.6, 55.6] & 55.5 [54.1, 57.0] & 54.8 [53.4, 56.3] \\
Qwen2.5-VL-7B-Instruct & 42.3 [41.1, 43.8] & 45.4 [44.3, 47.0] & 43.9 [42.9, 45.3] \\
InternVL3.5-4B & 44.5 [43.1, 46.1] & 40.4 [38.7, 42.3] & 42.5 [41.2, 43.9] \\
Qwen3-VL-4B-Instruct & 41.8 [40.6, 43.4] & 45.4 [44.1, 47.0] & 43.6 [42.4, 45.1] \\
Qwen2.5-VL-3B-Instruct & 38.0 [36.7, 39.4] & 40.8 [39.5, 42.4] & 39.4 [38.3, 40.8] \\
InternVL3.5-2B & 35.9 [34.7, 37.4] & 38.5 [37.3, 40.1] & 37.2 [36.2, 38.6] \\
Qwen3-VL-2B-Instruct & 36.7 [35.5, 38.2] & 40.0 [38.8, 41.6] & 38.4 [37.2, 39.8] \\
InternVL3-2B & 31.0 [29.8, 32.5] & 31.1 [29.9, 32.6] & 31.1 [30.0, 32.4] \\
\bottomrule
\end{tabular}
\endgroup
\end{table}

%% file: src/appendix-e-data_contamination_analysis.tex
\section{\rev{Supplementary Experimental Analysis}}
\label{app:additional_analysis}

\subsection{\rev{Domain Performance and Weighting}}
\label{app:topic_breakdown}

\rev{To assess how domain proportions affect model comparisons, we compute subtitle-condition accuracy for five representative models in Economy, Society, Science, Politics, and Culture, and compare question-weighted accuracy with an equally weighted mean across the five domains.}

\begin{table}[t!]
\centering
\caption{\rev{Per-domain exact-match accuracy (\%) with subtitles. Domain mean gives equal weight to the five domains.}}
\label{tab:domain-acc}
\begingroup
\ifshowrevisions\color{blue}\fi
\small
\setlength{\tabcolsep}{4pt}
\renewcommand{\arraystretch}{1.15}
\begin{tabular}{@{}lrrrrrrr@{}}
\toprule
Model & Economy & Society & Science & Politics & Culture & Avg Acc & Domain mean \\
\midrule
Gemini-3.1-Pro & \textbf{72.1} & 78.2 & \textbf{77.1} & \textbf{67.9} & \textbf{65.9} & \textbf{71.7} & \textbf{72.2} \\
GPT-5 & 69.2 & \textbf{78.8} & 71.9 & 62.8 & 63.1 & 68.8 & 69.2 \\
Claude-Sonnet-4.6 & 67.7 & 68.6 & 66.7 & 66.0 & 58.1 & 65.8 & 65.4 \\
Qwen3-VL-32B-Instruct & 61.9 & 67.3 & 57.3 & 62.8 & 47.5 & 59.8 & 59.4 \\
InternVL3.5-38B & 58.5 & 54.5 & 53.1 & 60.3 & 46.9 & 55.6 & 54.7 \\
\bottomrule
\end{tabular}
\endgroup
\end{table}

\rev{As shown in Table~\ref{tab:domain-acc}, equal domain weighting changes each model's accuracy by at most 0.9 percentage points and leaves the five-model ranking unchanged, indicating that their aggregate comparison is stable under this reweighting. Relative performance within individual domains can differ~\citep{dawn}: although Claude-Sonnet-4.6 has lower overall accuracy than GPT-5, it scores 66.0\% in Politics compared with GPT-5's 62.8\%.}

\subsection{\rev{Native Video Input Comparison}}
\label{app:native_video}

\rev{To examine how input format affects model performance, we compare Gemini-3.1-Pro and Gemini-3-Flash with 50 frames, 50 frames with subtitles, and complete audiovisual input. All three conditions include the question stems and answer options. The audiovisual condition retains the original audio track without additional subtitle text. Table~\ref{tab:native_video} reports the results.}

\begin{table}[t!]
\centering
\caption{\rev{Exact-match accuracy (\%) across input conditions. The frame conditions use the same 50 frames, without audio; audiovisual input includes the complete video and original audio.}}
\label{tab:native_video}
\begingroup
\ifshowrevisions\color{blue}\fi
\small
\begin{tabular*}{\linewidth}{@{\extracolsep{\fill}}llrrrr@{}}
\toprule
Model & Input & L1 & L2 & L3 & Overall \\
\midrule
Gemini-3.1-Pro & 50 frames & 78.9 & 66.4 & 55.1 & 68.2 \\
 & 50 frames + subtitles & 81.8 & 71.5 & 57.9 & 71.7 \\
 & Video + audio & 84.6 & 72.3 & 59.4 & 73.5 \\
\midrule
Gemini-3-Flash & 50 frames & 75.6 & 59.7 & 52.3 & 64.1 \\
 & 50 frames + subtitles & 78.0 & 68.8 & 55.1 & 68.5 \\
 & Video + audio & 80.3 & 66.1 & 55.9 & 68.9 \\
\bottomrule
\end{tabular*}
\endgroup
\end{table}

\rev{Compared with 50 frames alone, complete audiovisual input improves overall accuracy by 5.3 percentage points for Pro and 4.8 for Flash, with the largest gains on L2: 5.9 and 6.4 points, respectively. Inspection of sampled frames, source videos, and model responses links some improvements to briefly displayed evidence. For example, a basketball question asks for the total number of successful shots across two shooting styles, but the results table appears between two sampled frames. Pro answers incorrectly with frame input; with audiovisual input, it reads 18 and 14 successful shots and correctly sums them to 32. Similarly, a nuclear-power question asks about the initial state in 1960, but the sampled frames have already advanced to 1961. Both models answer with the wrong country count under frame input and correctly under audiovisual input. These cases show that sampling the relevant chart does not necessarily capture the values and temporal states required by the question.}

\paragraph{\rev{Subtitle supplementation.}}
\rev{What subtitles can recover depends on what the narration states. For example, the 50\% increase needed for a rainfall calculation is absent from the sampled frames but retained in the subtitles. With subtitles, Pro correctly computes 3,000 mm, matching its audiovisual response. In contrast, the basketball narration does not state the successful-shot counts, and Pro remains incorrect after subtitles are added. At the aggregate level, adding subtitles narrows the overall gap between frame-based and audiovisual input to 1.8 points for Pro and 0.4 for Flash.}

\paragraph{\rev{Reading available evidence.}}
\rev{Complete audiovisual input does not improve every response: relative to frames with subtitles, Pro gains 0.8 points on L2, whereas Flash loses 2.7 points. Our case inspection also identifies errors despite the relevant evidence being available. For example, an industry-wealth ranking chart is visible in both inputs. Both models correctly identify Fashion \& Retail as the second-ranked industry with frame input, but answer Manufacturing with audiovisual input. Here, the error concerns reading available chart evidence rather than missing it during sampling.}

\subsection{\rev{Test-Time Reasoning}}
\label{app:thinking}

\rev{The Qwen3-VL entries in Table~\ref{tab:benchmark} are Instruct variants. To examine test-time reasoning, we evaluate the corresponding Thinking variants of Qwen3-VL-8B and Qwen3-VL-32B under the same 50-frame input, with and without subtitles. Other settings follow Section~\ref{sec:experiment}, and the results are in Table~\ref{tab:thinking}.}

\begin{table*}[t!]
\centering
\caption{\rev{Exact-match accuracy (\%) of Qwen3-VL Instruct and Thinking variants.}}
\label{tab:thinking}
\begingroup
\ifshowrevisions\color{blue}\fi
\small
\setlength{\tabcolsep}{4pt}
\renewcommand{\arraystretch}{1.15}
\resizebox{\textwidth}{!}{%
\begin{tabular}{@{}l cc cc cc cc c@{}}
\toprule
 & \multicolumn{2}{c}{Level 1} & \multicolumn{2}{c}{Level 2} & \multicolumn{2}{c}{Level 3} & \multicolumn{2}{c}{Avg Acc} & \multirow{2}{*}{\textbf{Overall}} \\
\cmidrule(lr){2-3}\cmidrule(lr){4-5}\cmidrule(lr){6-7}\cmidrule(lr){8-9}
Model & w/o Sub. & w/ Sub. & w/o Sub. & w/ Sub. & w/o Sub. & w/ Sub. & w/o Sub. & w/ Sub. & \\
\midrule
Qwen3-VL-8B-Instruct & 59.0 & 64.7 & 28.5 & 33.9 & 38.0 & 38.9 & 44.1 & 48.2 & 46.2 \\
Qwen3-VL-8B-Thinking & 62.5 & 69.6 & 51.5 & 53.2 & 42.7 & 45.8 & 53.4 & 57.8 & 55.6 \\
\midrule
Qwen3-VL-32B-Instruct & 68.5 & 71.6 & 39.0 & 48.5 & 50.5 & 54.2 & 54.7 & 59.8 & 57.3 \\
Qwen3-VL-32B-Thinking & 72.3 & 77.4 & 55.3 & 64.4 & 48.0 & 50.2 & 60.2 & 65.5 & 62.9 \\
\bottomrule
\end{tabular}%
}
\endgroup
\end{table*}

\rev{Thinking raises overall accuracy at both scales, from 46.2\% to 55.6\% for 8B and from 57.3\% to 62.9\% for 32B, though the gains are uneven across levels. L2 improves the most: without and with subtitles, 8B rises from 28.5\% and 33.9\% to 51.5\% and 53.2\%, and 32B from 39.0\% and 48.5\% to 55.3\% and 64.4\%. L1 also gains 3.5 to 5.8 percentage points. L3 changes inconsistently: 8B gains 4.7 and 6.9 points, whereas 32B drops from 50.5\% and 54.2\% to 48.0\% and 50.2\%.}

\rev{{As with subtitles (Section~\ref{sec:input_evidence}), Thinking also yields its largest gain on L2.} The drop for the 32B model on L3 indicates that additional reasoning time does not necessarily produce a more accurate narrative judgment. Even with Thinking enabled, the strongest open-source result, 65.5\% for Qwen3-VL-32B with subtitles, remains below Gemini-3.1-Pro and GPT-5 under the same condition (71.7\% and 68.8\%) and well below the human score of 91.7\%.}

\subsection{\rev{Higher Frame Budgets}}
\label{app:frame100}

\rev{To examine performance beyond 50 frames, we extend the inputs of Qwen3-VL-32B and InternVL3.5-38B to 100 frames. Commercial APIs limit the number of images per request, so this comparison focuses on open-source models. Both budgets include subtitles. The 100-frame input retains the original 50 frames and adds samples between them; Table~\ref{tab:frame100} compares the two budgets.}

\begin{table}[t!]
\centering
\caption{\rev{Exact-match accuracy (\%) at 50 and 100 frames with subtitles.}}
\label{tab:frame100}
\begingroup
\ifshowrevisions\color{blue}\fi
\small
\setlength{\tabcolsep}{5pt}
\renewcommand{\arraystretch}{1.15}
\begin{tabular}{@{}llrrrr@{}}
\toprule
Model & Frames & Overall & L1 & L2 & L3 \\
\midrule
Qwen3-VL-32B & 50 & 59.8 & 71.6 & 48.5 & 54.2 \\
 & 100 & 62.4 & 76.3 & 50.8 & 54.2 \\
 & Change & +2.6 & +4.7 & +2.3 & 0.0 \\
\midrule
InternVL3.5-38B & 50 & 55.6 & 65.6 & 48.1 & 48.9 \\
 & 100 & 55.0 & 67.0 & 46.1 & 46.7 \\
 & Change & $-$0.6 & +1.4 & $-$2.0 & $-$2.2 \\
\bottomrule
\end{tabular}
\endgroup
\end{table}

\rev{Increasing the budget from 50 to 100 frames raises Qwen's overall accuracy by 2.6 percentage points and lowers InternVL's by 0.6. L1 improves for both models, whereas L2 and L3 do not move together: Qwen's L3 stays at 54.2\%, while InternVL declines on L2 and L3. This extends the trend in Section~\ref{sec:frame_budget}. Denser sampling can add visual detail for basic reading, but the gain does not reliably carry over to temporal reasoning or narrative understanding.}

\subsection{\rev{Multi-Select Diagnostics}}
\label{app:multiselect_diagnostics}

\paragraph{\rev{Evaluation.}}
\rev{To distinguish option identification from complete answers on multi-select questions, we report option-level precision, recall, and F1 alongside exact match. The statistics cover Causal Reasoning, Argument Synthesis, {Narrative Structure}, Visual Communication Intent, and Counterfactual Analysis. We compare Gemini-3.1-Pro, GPT-5, Claude-Sonnet-4.6, Qwen3-VL-32B-Instruct, and InternVL3.5-38B under both subtitle conditions.}

\rev{We compute each metric per question, average within each model, and then average equally across the five models.}

\begin{table}[t!]
\centering
\caption{\rev{Multi-select performance (\%). Exact match is computed on multi-select items only. Gold and Selected denote the mean numbers of gold and predicted options. Empty or invalid predictions receive zero scores.}}
\label{tab:multiselect_metrics}
\begingroup
\ifshowrevisions\color{blue}\fi
\small
\begin{tabular*}{\linewidth}{@{\extracolsep{\fill}}lrrrrrr@{}}
\toprule
Task & EM & Precision & Recall & F1 & Gold & Selected \\
\midrule
\multicolumn{7}{l}{\textit{Without Subtitles}} \\
Causal Reasoning & 27.9 & 94.3 & 71.8 & 79.9 & 2.7 & 2.1 \\
Argument Synthesis & 37.3 & 93.5 & 73.8 & 80.6 & 3.0 & 2.3 \\
{Narrative Structure} & 42.7 & 92.2 & 78.3 & 82.7 & 2.4 & 2.1 \\
Visual Communication Intent & 53.9 & 95.0 & 81.0 & 85.3 & 2.2 & 1.9 \\
Counterfactual Analysis & 62.6 & 93.1 & 86.6 & 88.1 & 2.3 & 2.1 \\
\midrule
\multicolumn{7}{l}{\textit{With Subtitles}} \\
Causal Reasoning & 37.5 & 94.1 & 75.7 & 82.2 & 2.7 & 2.2 \\
Argument Synthesis & 53.9 & 96.0 & 83.8 & 88.0 & 3.0 & 2.6 \\
{Narrative Structure} & 49.0 & 93.8 & 82.4 & 86.0 & 2.4 & 2.1 \\
Visual Communication Intent & 54.6 & 95.4 & 81.4 & 85.8 & 2.2 & 1.9 \\
Counterfactual Analysis & 61.7 & 94.2 & 86.3 & 88.4 & 2.3 & 2.1 \\
\bottomrule
\end{tabular*}
\endgroup
\end{table}

\paragraph{\rev{Answer completeness.}}
\rev{Table~\ref{tab:multiselect_metrics} shows incomplete coverage of correct options across all five multi-select types: under both input conditions, precision exceeds recall, and models select fewer options on average than the gold answers contain. With subtitles, Argument Synthesis reaches 96.0\% precision but 83.8\% recall and 53.9\% exact match. Models identify most of the correct content, but often fail to include all required options. Improving these tasks therefore calls for more complete coverage of the answer set while maintaining accurate option selection.}

\paragraph{\rev{Task-specific performance.}}
\rev{Because the number of required options varies across task types, we further group questions by gold answer-set size. As shown in Table~\ref{tab:multiselect_gold_size}, with subtitles, Causal Reasoning has F1 scores of 74.1\% and 85.3\% in the two- and three-option groups, respectively, lower than the other four task types. Argument Synthesis reaches 90.1\% F1 in the three-option group, second only to Counterfactual Analysis, but its exact-match accuracy is 57.7\%. These tasks highlight different priorities for improvement: Causal Reasoning requires better identification and coverage of supporting options, whereas Argument Synthesis often yields many correct selections without a complete answer set.}

\begin{table}[t!]
\centering
\caption{\rev{Multi-select EM and F1 (\%) by the number of correct options, using the same aggregation as Table~\ref{tab:multiselect_metrics}.}}
\label{tab:multiselect_gold_size}
\begingroup
\ifshowrevisions\color{blue}\fi
\small
\begin{tabular*}{\linewidth}{@{\extracolsep{\fill}}lrrrr@{}}
\toprule
 & \multicolumn{2}{c}{Two Correct Options} & \multicolumn{2}{c}{Three Correct Options} \\
\cmidrule(lr){2-3}\cmidrule(l){4-5}
Task & EM & F1 & EM & F1 \\
\midrule
\multicolumn{5}{l}{\textit{Without Subtitles}} \\
Causal Reasoning & 40.0 & 74.8 & 23.4 & 81.8 \\
Argument Synthesis & 37.1 & 69.8 & 40.3 & 83.9 \\
{Narrative Structure} & 50.4 & 81.7 & 32.9 & 84.1 \\
Visual Communication Intent & 58.1 & 85.3 & 38.1 & 84.2 \\
Counterfactual Analysis & 61.9 & 87.1 & 65.0 & 91.1 \\
\midrule
\multicolumn{5}{l}{\textit{With Subtitles}} \\
Causal Reasoning & 36.9 & 74.1 & 37.7 & 85.3 \\
Argument Synthesis & 54.3 & 83.0 & 57.7 & 90.1 \\
{Narrative Structure} & 53.3 & 84.1 & 43.4 & 88.4 \\
Visual Communication Intent & 57.6 & 85.3 & 42.4 & 86.3 \\
Counterfactual Analysis & 61.3 & 87.2 & 63.2 & 91.6 \\
\bottomrule
\end{tabular*}
\endgroup
\end{table}

\rev{The remaining groups contain one Visual Communication Intent question with a single correct option, for which EM and F1 are both 100\% under both conditions, and five Argument Synthesis questions with four correct options. For the latter, EM is 20.0\% without subtitles and 32.0\% with subtitles; F1 is 76.5\% and 83.4\%, respectively.}

\subsection{\rev{Video Duration Analysis}}
\label{app:duration_analysis}

\rev{To examine the relationship between video duration and model performance, we group evaluation videos into four duration intervals: at most 3 minutes, 3--6 minutes, 6--9 minutes, and more than 9 minutes. Table~\ref{tab:duration_bin} reports accuracy for all 19 models under the \emph{w/o sub} and \emph{w/ sub} conditions.}

\paragraph{\rev{Duration-related declines without subtitles.}}
\rev{Models exhibit different patterns of performance decline on longer videos. GPT-5 achieves approximately 68\% accuracy in the first three duration groups, dropping to 59.4\% in the longest group. Claude-Sonnet-4.6 shows a gradual decline from 64.1\% in the shortest group to 59.0\% in the longest. Among open-source models, Qwen3-VL-32B and InternVL3.5-38B score 6.4 and 5.8 percentage points lower on the longest videos than on the shortest. GPT-4o, by comparison, stays between 54.8\% and 57.8\% across the four groups. These results distinguish accuracy from consistency across durations: GPT-5 outperforms GPT-4o in every group while showing a larger short-to-long gap.}

\paragraph{\rev{Subtitles narrow short-to-long performance gaps.}}
\rev{With subtitles, GPT-5's accuracy on the longest videos rises from 59.4\% to 66.9\%, narrowing its short-to-long gap from 8.9 to 1.4 percentage points. Qwen3-VL-32B improves from 53.1\% to 60.4\% on the longest videos, approaching its 60.7\% accuracy on the shortest; Qwen3-VL-8B's gap similarly shrinks from 6.0 to 0.9 points. Claude-Sonnet-4.6's gap decreases from 5.1 to 1.2 points. This pattern across closed- and open-source models shows that subtitle information helps these models maintain performance on longer data videos. Gemini-3.1-Pro's gains are concentrated in the middle duration groups: 4.1 and 9.2 points for 3--6 and 6--9 minutes, respectively, compared with 0.3 points for videos longer than 9 minutes.}

\begin{table}[t!]
\centering
\caption{\rev{Model accuracy (\%) across four video-duration intervals, reported separately under \emph{w/o sub} and \emph{w/ sub}.}}
\label{tab:duration_bin}
\small
\setlength{\tabcolsep}{4pt}
\resizebox{\textwidth}{!}{
\begin{tabular}{l cccc cccc}
\toprule
 & \multicolumn{4}{c}{\emph{w/o sub}} & \multicolumn{4}{c}{\emph{w/ sub}} \\
\cmidrule(lr){2-5} \cmidrule(lr){6-9}
Model & $\leq$3\,min & 3--6\,min & 6--9\,min & $>$9\,min & $\leq$3\,min & 3--6\,min & 6--9\,min & $>$9\,min \\
\midrule
\multicolumn{9}{c}{\textit{Closed-Source Models}} \\
\midrule
Gemini-3.1-Pro & 70.2 & 70.8 & 64.5 & 66.7 & 71.4 & 74.9 & 73.7 & 67.0 \\
GPT-5 & 68.3 & 68.6 & 67.9 & 59.4 & 68.3 & 72.4 & 67.0 & 66.9 \\
Gemini-3-Flash & 64.1 & 68.3 & 60.4 & 62.6 & 65.3 & 72.1 & 69.6 & 67.0 \\
GPT-4o & 57.1 & 57.8 & 54.8 & 54.9 & 54.4 & 58.4 & 59.9 & 59.7 \\
Claude-Sonnet-4.6 & 64.1 & 62.2 & 59.9 & 59.0 & 64.9 & 67.6 & 67.3 & 63.7 \\
\midrule\multicolumn{9}{c}{\textit{Open-Source Models}}\\\midrule
Qwen3-VL-32B & 59.5 & 54.6 & 52.1 & 53.1 & 60.7 & 60.0 & 58.5 & 60.4 \\
InternVL3.5-38B & 54.2 & 52.7 & 56.7 & 48.4 & 55.0 & 57.8 & 59.0 & 51.6 \\
InternVL3.5-14B & 46.2 & 48.3 & 45.2 & 41.4 & 47.7 & 54.3 & 50.7 & 45.8 \\
InternVL3.5-8B & 47.3 & 45.7 & 45.2 & 41.0 & 47.3 & 48.9 & 49.3 & 44.7 \\
Qwen3-VL-8B & 49.2 & 41.6 & 43.3 & 43.2 & 49.6 & 47.9 & 47.5 & 48.7 \\
InternVL3-8B & 49.8 & 40.0 & 45.6 & 40.1 & 50.2 & 44.8 & 51.6 & 42.3 \\
Qwen2.5-VL-32B & 57.3 & 55.9 & 47.9 & 53.8 & 56.5 & 56.8 & 55.8 & 53.1 \\
Qwen2.5-VL-7B & 44.7 & 44.1 & 40.6 & 39.9 & 45.4 & 46.0 & 48.4 & 43.2 \\
InternVL3.5-4B & 44.3 & 46.3 & 45.2 & 42.5 & 43.5 & 39.7 & 40.1 & 38.8 \\
Qwen3-VL-4B & 45.4 & 45.1 & 39.2 & 37.4 & 46.9 & 47.0 & 44.2 & 43.6 \\
Qwen2.5-VL-3B & 38.9 & 38.4 & 38.2 & 36.6 & 38.2 & 41.0 & 46.1 & 39.6 \\
InternVL3.5-2B & 38.2 & 37.5 & 33.6 & 34.1 & 41.6 & 40.6 & 37.8 & 34.4 \\
Qwen3-VL-2B & 37.8 & 38.1 & 37.8 & 33.7 & 41.2 & 39.4 & 42.9 & 38.1 \\
InternVL3-2B & 32.4 & 29.7 & 31.3 & 31.5 & 31.2 & 30.0 & 34.1 & 30.4 \\
\bottomrule
\end{tabular}
}
\end{table}

\subsection{\rev{Representative Error Cases}}
\label{app:gemini_error_cases}

\rev{To complement the error-type analysis of Gemini-3.1-Pro in Section~\ref{sec:error_analysis}, we present two representative failures in the model's actual responses. In both cases the model reads the numbers printed on the chart correctly. The failure is that it does not follow a constraint in the question: one shifts the specified time range, and the other ignores the requirement that the evidence come from the chart.}

\begin{figure*}[t]
  \centering
  \includegraphics[width=\textwidth]{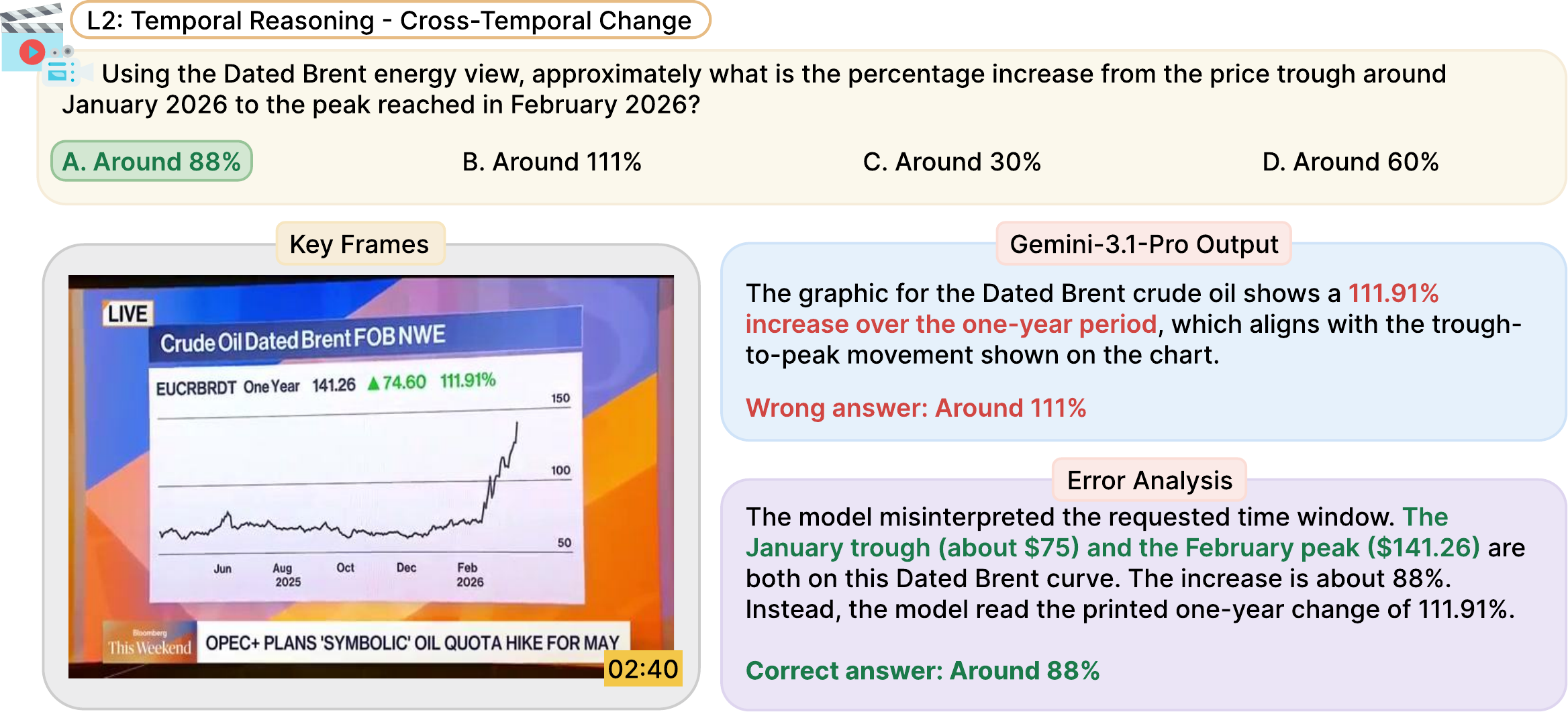}
  \caption{\rev{An instruction-understanding error of Gemini-3.1-Pro. The model is distracted by on-chart text and substitutes the printed one-year change for the requested time-window increase.}}
  \label{fig:gemini_error_time_window}
\end{figure*}

\rev{Figure~\ref{fig:gemini_error_time_window} is the time-range case. The question asks for the increase from the trough around January 2026 to the peak in February 2026. The model locates the relevant chart, but it is distracted by the more salient one-year increase of 111.91\% printed on the chart and answers with that value, rather than computing the trough-to-peak gain over the requested interval (approximately 88\%). With multimodal input, the model can be drawn to salient on-chart text and pass over an explicit time window in the question.}

\begin{figure*}[t]
  \centering
  \includegraphics[width=\textwidth]{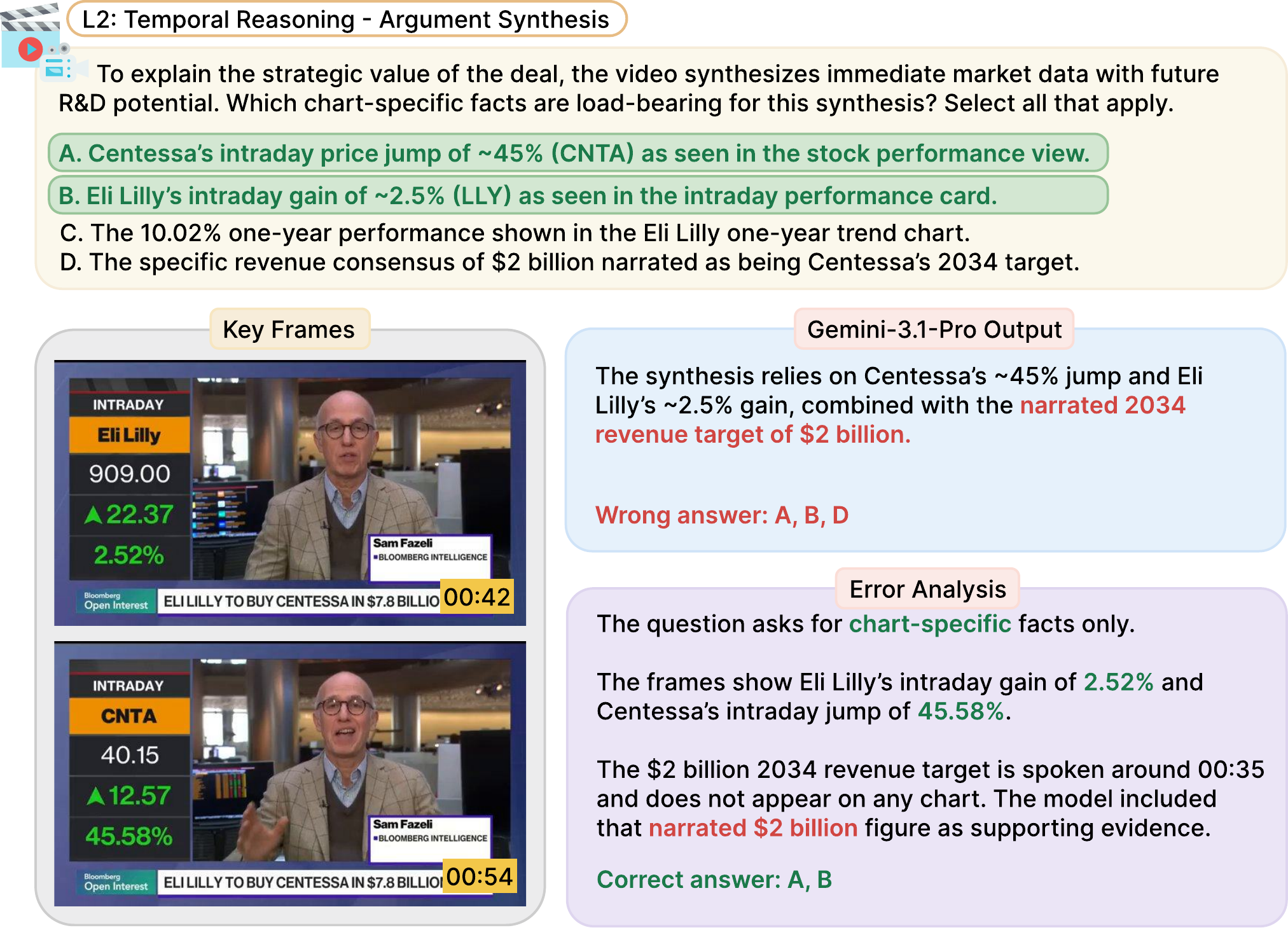}
  \caption{\rev{An instruction-understanding error of Gemini-3.1-Pro. The question admits only facts shown on the chart, yet the model still treats the narrated 2034 revenue target of \$2 billion as supporting evidence.}}
  \label{fig:gemini_error_evidence_source}
\end{figure*}

\rev{Figure~\ref{fig:gemini_error_evidence_source} is the evidence-source case. The question asks for chart-specific facts that bear on the strategic value of the deal. The frames show Eli Lilly's intraday gain of 2.52\% at 00:42 and Centessa's (CNTA) intraday jump of 45.58\% at 00:54, both printed on the performance cards. The 2034 revenue target of about \$2 billion is only spoken around 00:35 and does not appear on any chart. The model treats this narrated figure, together with the two on-chart gains, as supporting evidence and answers A, B, D; the correct answer is A, B. The model can read the chart correctly and still place a number from the narration into an answer that admits only chart facts.}

\subsection{\rev{Diagnostic Analysis of an Open-Source Model}}
\label{app:qwen_error_analysis}

\rev{We manually inspect 120 randomly sampled incorrect responses from a representative open-source model, Qwen3-VL-32B, with 50-frame input and subtitles. The three categories identified in Section~\ref{sec:error_analysis} are also prominent for this model, though they take somewhat different forms.}

\paragraph{\rev{Perceptual errors.}}
\rev{At the basic visual extraction stage, the main bottleneck for the open-source model lies in misreading temporal and structural information. (1) Temporal anchor confusion: when computing the decline in a brand's sales, the model reads the historical peak value shown in the chart (from 2016) as the current-year figure (2024), causing the subsequent derivation to deviate entirely. (2) Chart structure misjudgment: the model often fails to correctly identify the chart's axis configuration. For example, given a bar chart whose $y$-axis clearly starts at 0\%, the model nonetheless claims in its reasoning text that the chart uses a truncated $y$-axis to exaggerate the visual difference.}

\paragraph{\rev{Reasoning errors.}}
\rev{Most errors that can be localized at the reasoning stage occur when linking facts to a claim. On causal-reasoning and narrative-argument questions, the model typically extracts the correct values and trends from the relevant frames; however, when judging whether these visual facts support a given option's claim, it frequently reaches the opposite conclusion, failing to translate locally correct facts into valid support for the overall option.}

\paragraph{\rev{Instruction and mapping errors.}}
\rev{This category is particularly prominent in the open-source sample and mostly reflects a mismatch between the model's internal reasoning and its final output. (1) Option-mapping inconsistency: the model reaches the correct conclusion during its textual reasoning but disconnects from it when producing the final predicted letter. For example, when comparing per-capita GDP between two countries, the model correctly computes a roughly 20-fold difference and states in its rationale that this matches option C, yet its final prediction field contains D. (2) Constraint-scope drift: when a question explicitly restricts the comparison to a specific year, the model disregards this temporal constraint during extraction and mixes in projected or other-year values from the same chart.}

\rev{Overall, although the open-source model demonstrates some competence in extracting data from individual frames, it continues to face challenges in integrating information into a coherent overall picture and maintaining consistency over long videos. Both the difficulty of linking facts to claims and the disconnect between correct internal derivations and final outputs point to the importance of maintaining cross-frame coherence and closely following instructions over long contexts, and suggest exploring candidate-answer verification and selection~\citep{alpha-sql,deepeye-sql,dpc,lin2025lead}.}

\subsection{\rev{Upload-Time and View-Count Analysis}}
\label{app:contamination}

\rev{Publicly available data videos may be included in model pretraining corpora and thereby affect evaluation results. To examine potential data contamination, we group videos by upload time and view count and compare the overall accuracy of the 19 models evaluated in Section~\ref{sec:experiment} across groups, examining whether older or more-viewed videos show a consistent performance advantage.}

\paragraph{\rev{Upload time.}}
\rev{We first examine whether model performance varies with video upload time. We divide videos into three groups by YouTube upload date: before 2025 (214 videos), within 2025 (399 videos), and on or after 2026-01-01 (348 videos). The results show no consistent accuracy advantage for older videos. Gemini-3.1-Pro achieves 68.9\%, 68.8\%, and 71.9\% across the three groups, all close to its overall accuracy of 70.0\% in Table~\ref{tab:benchmark}; GPT-4o achieves 56.1\%, 57.3\%, and 57.5\%, and InternVL3-8B achieves 45.1\%, 44.1\%, and 46.5\%. These three models perform similarly across year groups and attain their highest accuracy on the most recent group. The difference is larger for GPT-5: accuracy on videos published in 2026 is 69.9\%, exceeding its 63.2\% on pre-2025 videos by 6.7 percentage points. Overall, earlier publication does not correspond to higher accuracy, and models maintain comparable or better performance on recent videos.}

\paragraph{\rev{View count.}}
\rev{We next examine whether model performance improves with video view count. We divide the 961 videos into four equal-frequency groups by view count, from the lowest (Group~1) to the highest (Group~4). Accuracy does not increase monotonically with view count. GPT-5 achieves 69.1\%, 70.7\%, 68.4\%, and 60.9\% across the four groups; Gemini-3.1-Pro achieves 70.7\%, 75.7\%, 70.6\%, and 62.2\%; and Qwen3-VL-32B achieves 53.3\%, 57.8\%, 64.0\%, and 54.4\%. GPT-5 and Gemini-3.1-Pro both attain their highest accuracy in Group~2 and their lowest in the most-viewed group. Qwen3-VL-32B performs best in Group~3, with a 9.6-percentage-point decline in Group~4. These results show no consistent performance advantage for highly viewed videos: more widely viewed content is not necessarily easier for models to answer correctly.}

\rev{Neither analysis shows a systematic performance advantage for older or more widely viewed content.}

%% file: src/appendix-f-limitation.tex
\section{Limitations and Future Work}
\label{app:limitations_future}

\rev{\benchmark has the following limitations: (1) As this work focuses on the foundational capability of data video comprehension, data video generation and editing are not yet covered. (2) To ensure objective, standardized, and reproducible evaluation at scale, the benchmark currently relies on multiple-choice questions and does not yet include open-ended subjective items. (3) The benchmark currently consists primarily of English data videos, which limits the evaluation of multilingual data video comprehension.}

\rev{Future work will extend the evaluation to data video generation and editing, introduce open-ended items along with automated scoring schemes, and gradually expand to multilingual datasets. Building on the rich structured annotations already collected, we plan to construct a fine-tuning dataset to train models for data video understanding. Furthermore, to foster a dynamic research ecosystem, we will establish and maintain an official evaluation website and public leaderboard. This platform will periodically incorporate newly released data videos to expand the benchmark, continuously evaluate the latest models, and support researchers in submitting their custom models for evaluation, thereby driving long-term progress in the field.}